%% file: main.tex
\PassOptionsToPackage{dvipsnames,table}{xcolor}
\documentclass[safefonts]{fairmeta}
\usepackage{amsmath}
\usepackage{amssymb}
\usepackage{amsfonts}
\usepackage{amsthm}
\usepackage{nicefrac}
\usepackage{tikz}
\usepackage{subfigure}
\usepackage{enumitem}
\usepackage{rotating}
\usepackage{adjustbox}
\usepackage{colortbl}
\usepackage[normalem]{ulem}
\usepackage{listings}
\usepackage{pifont}
\usepackage{wrapfig}

\definecolor{mydarkblue}{HTML}{1A4F8B}
\definecolor{metabg}{HTML}{F8FAFD}
\definecolor{metafg}{HTML}{202124}
\definecolor{metablue}{HTML}{1A73E8}
\hypersetup{
  colorlinks=true,
  linkcolor=mydarkblue,
  citecolor=mydarkblue,
  urlcolor=mydarkblue,
  filecolor=mydarkblue
}

\definecolor{TableHeaderLavender}{RGB}{235,232,243}
\definecolor{TableStripeSoft}{RGB}{241,241,241}
\definecolor{TableStripeGreen}{rgb}{.961,.985,.970}
\definecolor{TableStripePurple}{RGB}{239,233,250}
\definecolor{grey}{rgb}{0.89,0.71,0.57}
\definecolor{pink}{rgb}{1,0.94,1}
\definecolor{purple}{rgb}{0.84,0.78,1}
\definecolor{white}{rgb}{1,1,1}
\definecolor{backred}{RGB}{255,190,190}
\definecolor{backblue}{RGB}{210,230,250}
\definecolor{mygrey}{RGB}{200,200,200}
\definecolor{codegreen}{rgb}{0,0.6,0}
\definecolor{codegray}{rgb}{0.5,0.5,0.5}
\definecolor{codepurple}{rgb}{0.58,0,0.82}
\definecolor{backcolour}{rgb}{0.95,0.95,0.92}
\definecolor{lightyellow}{RGB}{255,252,51}
\definecolor{lightgreen}{RGB}{204,255,204}
\definecolor{DarkGreen}{RGB}{0,100,0}
\definecolor{DarkYellow}{rgb}{0.8,0.8,0.0}
\definecolor{DarkBrown}{rgb}{0.4,0.2,0.1}
\definecolor{DarkBlue}{rgb}{0.0,0.0,0.5}
\definecolor{DarkRed}{rgb}{0.5,0.0,0.0}

\lstdefinestyle{mystyle}{
  backgroundcolor=\color{backcolour},
  commentstyle=\color{codegreen},
  keywordstyle=\color{magenta},
  numberstyle=\tiny\color{codegray},
  stringstyle=\color{codepurple},
  basicstyle=\ttfamily\footnotesize,
  breakatwhitespace=false,
  breaklines=true,
  captionpos=b,
  keepspaces=true,
  numbers=left,
  numbersep=5pt,
  showspaces=false,
  showstringspaces=false,
  showtabs=false,
  tabsize=2
}
\newcommand{\githubicon}{\raisebox{-1.5pt}{\includegraphics[height=1.03em]{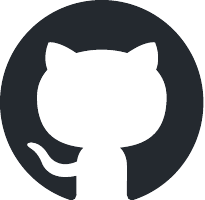}}}
\newcommand{\huggingfaceicon}{\raisebox{-1.5pt}{\includegraphics[height=0.96em]{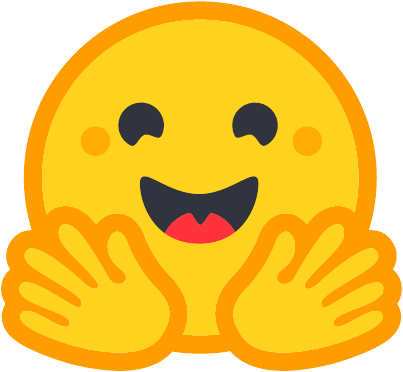}}}

\newcommand{\method}{\textsc{VideoGAIA}}
\newcommand{\modelicon}[2]{\raisebox{-0.2\height}{\includegraphics[height=#1]{icons/#2}}}
\renewcommand{\cite}{\citep}

\title{VideoGAIA: A Benchmark for General AI Assistants on Agentic Video Understanding}

\author[*,1]{Fan Zhang}
\author[*,2]{Guangming Yao}
\author[3]{Jinyang Wu}
\author[3]{Hao Wu}
\author[4]{Zheng Lian}
\author[5]{Xinyu Geng}
\author[2]{Jingdong Chen}
\author[\dagger,2]{Yi Yuan}
\author[\dagger,1]{Pheng-Ann Heng}

\affiliation[1]{The Chinese University of Hong Kong}
\affiliation[2]{Ant Group}
\affiliation[3]{Tsinghua University}
\affiliation[4]{Tongji University}
\affiliation[5]{The Hong Kong University of Science and Technology}

\contribution[*]{Equal contribution}
\contribution[\dagger]{Corresponding authors.}
\metadata[\githubicon~Code]{\url{https://github.com/zfkarl/VideoGAIA}}
\metadata[\huggingfaceicon~Data]{\url{https://huggingface.co/datasets/Karl28/VideoGAIA}}
\abstract{

\input{1_abstract}}

\begin{document}

\maketitle

\input{2_introduction}

\input{3_related_work}
\input{4_method}

\input{5_experiment}

\input{6_conclusion}

\bibliographystyle{assets/plainnat}
\bibliography{rec}

\input{7_appendix}

\end{document}

%% file: 1_abstract.tex
Video understanding is a fundamental task for evaluating the capabilities of multimodal large language models (MLLMs). However, existing leading models have already achieved approximately 90\% accuracy on the Video-MME leaderboard, suggesting that conventional single-turn video understanding tasks are becoming increasingly saturated and insufficient for assessing the intelligence of advanced MLLMs. Towards this end, we introduce \textbf{VideoGAIA}, an agentic \underline{\textbf{Video}} understanding benchmark for \underline{\textbf{G}}eneral artificial intelligence (\underline{\textbf{AI}}) \underline{\textbf{A}}ssistants. Moving beyond one-shot video question answering, VideoGAIA formulates video understanding as a multi-turn, tool-augmented interaction process, where models must iteratively perceive videos, invoke external tools, gather complementary information, and integrate multimodal evidence across turns. VideoGAIA contains 271 model-human co-designed tasks covering diverse and complex real-world scenarios. Each video-question-answer instance is independently verified by three human experts to ensure both correctness and appropriate difficulty. All evaluated MLLMs, including frontier models such as GPT-5.5 and Kimi-K3, achieve less than 60\% accuracy on VideoGAIA, highlighting its value as a high-quality and timely benchmark for evaluating next-generation MLLMs. We hope that VideoGAIA will facilitate the transition from conventional video understanding toward \textbf{agentic video understanding}.
%Code is available at \url{https://github.com/zfkarl/VideoGAIA}.

%% file: 2_introduction.tex
\section{Introduction}

Video understanding is a fundamental task in multimodal understanding and has long served as a foundation for evaluating multimodal large language models (MLLMs). Over the past several years, numerous benchmarks \cite{wu2024longvideobench,zhao2025mmvu,li2024mvbench} have been proposed to assess model capabilities in multimodal perception, temporal reasoning, and object recognition.

\begin{figure*}[t]
    \centering
    \includegraphics[width=\linewidth]{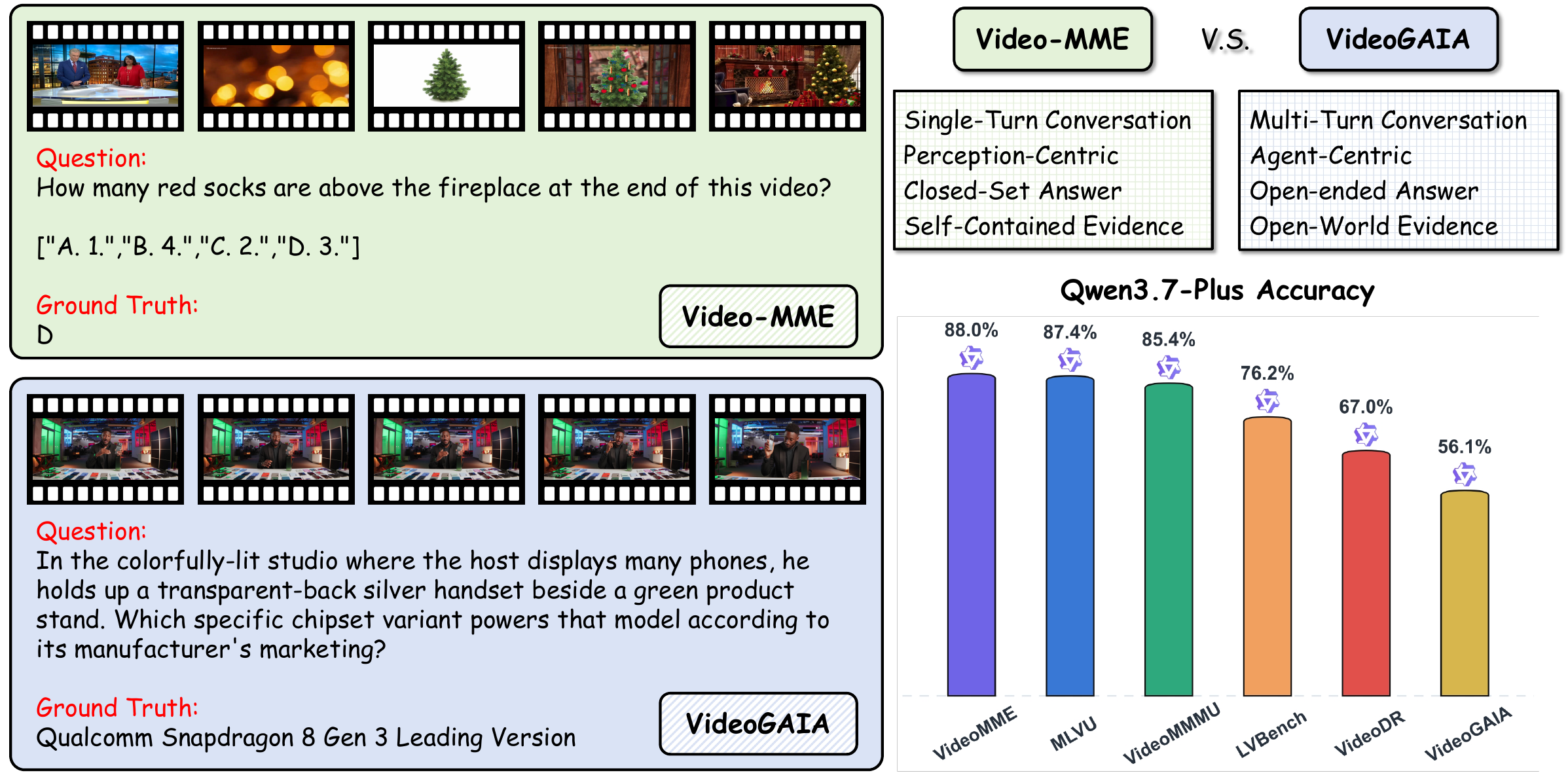}
    \caption{VideoGAIA focuses on agentic video understanding, which is more challenging compared with conventional video understanding benchmarks.}
    \label{fig:intro}
\end{figure*}

Despite their success, the rapid progress of foundation models has driven the performance of existing video understanding benchmarks close to saturation. On widely adopted benchmarks such as Video-MME \cite{fu2025video}, VideoMMMU \cite{hu2026video}, and MLVU \cite{zhou2024mlvu}, the strongest MLLMs \cite{qwen37plus,openaigpt55,gemini31pro} now achieve accuracies approaching 90\%, suggesting that conventional single-turn video question answering has become a nearly solved problem. This trend calls for a new evaluation paradigm that better reflects the capabilities required by next-generation multimodal systems.
Meanwhile, the emergence of multimodal agents is reshaping how artificial intelligence (AI) systems interact with the world, yet the development and evaluation of video agents remain largely underexplored. Towards this end, we introduce \textbf{VideoGAIA}, an agentic video understanding benchmark for general AI assistants. Going beyond traditional single-turn video understanding, VideoGAIA provides a comprehensive evaluation suite that measures the ability of video agents to iteratively perceive videos, engage in multi-turn interactions, invoke external tools, and integrate information from multiple sources to solve complex real-world tasks (shown in Figure \ref{fig:intro}). Even with access to additional tools, the frontier model Qwen3.7-Plus \cite{qwen37plus} still achieves substantially lower accuracy on VideoGAIA than on other video understanding benchmarks.

To design a video agent benchmark that addresses the most pressing needs of the community, we first ask a fundamental question: \textit{what are contemporary MLLMs already good at, and where do they still fall short?} On the one hand, continued scaling in model size and training data has endowed frontier models with strong general capabilities and extensive world knowledge \cite{qwen3max,kimik3}. They have become highly proficient in foundational tasks such as instruction following, visual recognition, and temporal understanding, which largely explains the saturation of single-turn video understanding benchmarks. On the other hand, foundation models are undergoing a broader transition from conversational chatbots to general-purpose productivity tools \cite{openaigpt55,glm52,luo2025large,zhang2026orchestra}. Capabilities essential to this transition, including tool use, long-horizon reasoning, information aggregation, and context management, remain under active development and are far from fully mature.
Video understanding should therefore evolve accordingly for the agentic era. Its evaluation should no longer be limited to asking a model to watch a video and answer a single question passively. Instead, a capable video agent should jointly interpret textual instructions and visual-temporal evidence, identify information that cannot be obtained from the video alone, proactively use available tools to seek complementary information, and efficiently synthesize the collected evidence across multiple interaction steps. As shown in the left part of Figure \ref{fig:intro}, the ultimate goal of a video benchmark is not merely to test whether a model can answer questions about a video, but whether it can use videos as a source of evidence to accomplish concrete, real-world tasks. This perspective forms the central design principle of VideoGAIA and motivates our shift from conventional video question answering toward agentic video understanding.

The construction of VideoGAIA follows a multi-stage human-model co-design pipeline. We begin with a pool of 100K publicly available videos collected from the Internet and employ frontier models, including Gemini-3.1-Pro \cite{gemini31pro} and GPT-5.5 \cite{openaigpt55}, to support annotation and quality-control stages. These stages include frame-level video captioning, evidence-grounded task generation, rubric-based quality control, cross-modal filtering, tool-necessity assessment, and human crowdsourced annotation. Each model-filtered task is independently reviewed by at least three human experts. Only tasks that receive unanimous approval, or majority approval followed by an additional final review, are retained. The resulting task pool spans six task categories, including Culture, Geography, Daily Life, History, Technology, and Economics, and is further divided into two difficulty levels, L1 and L2. After more than 100 hours of annotation by 15 annotators, this process ultimately yields 271 high-quality video-agent tasks that constitute the VideoGAIA test set.

\begin{figure*}[t]
    \centering
    \includegraphics[width=\linewidth]{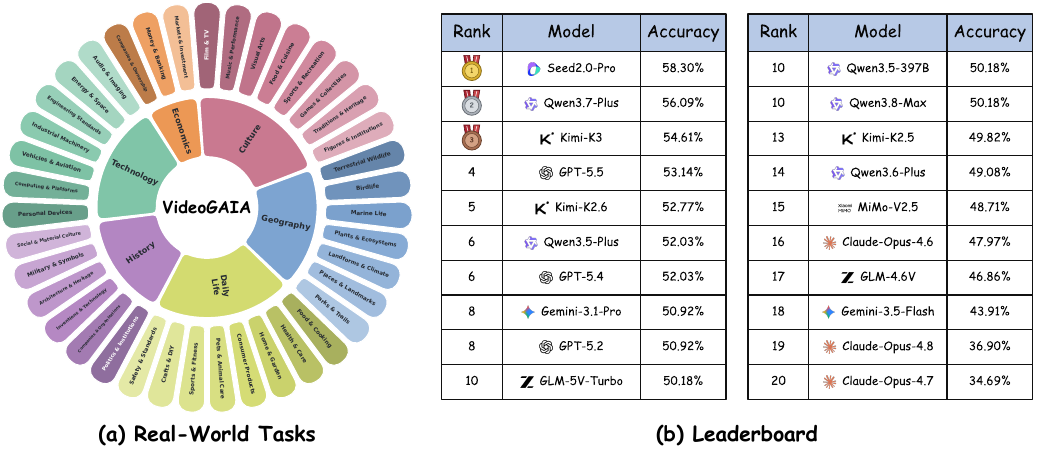}
    \caption{\textbf{(a) Real-World Tasks.} VideoGAIA spans 6 key domains and 39 subtasks. \textbf{(b) Leaderboard}. Performance comparison of 20 frontier MLLMs is shown under the agent loop setting.}
    \label{fig:stat}
\end{figure*}

Following data curation, we conduct a comprehensive evaluation of 20 cutting-edge frontier MLLMs. To ensure a fair comparison, all models are evaluated under the same experimental setting using a ReAct-style \cite{yao2022react} agent loop equipped with three tools: web search for retrieving external information, page visit for converting webpages into textual content, and thinking with videos for revisiting and inspecting relevant video segments. The results show that all evaluated models achieve accuracies between 30\% and 60\%. Even the strongest model Seed2.0-Pro \cite{seed2026seed2} reaches only 58.30\% accuracy, substantially lower than the performance on conventional video understanding benchmarks.
These results in the leaderboard (as shown in Figure \ref{fig:stat}) indicate that agentic video understanding remains far from solved and presents substantial opportunities for future research. We hope it will encourage further research exploration of model architectures, training strategies, and agent harnesses, ultimately enabling video agents to genuinely understand videos, assist users in completing real-world tasks, and deliver meaningful productivity gains.

To summarize, the main contributions of this work are threefold:
\begin{itemize}[leftmargin=*]
\item \textbf{Paradigm Shift.} We advocate a transition from single-turn video understanding to agentic video understanding. This new paradigm encourages video agents to move beyond answering questions about videos and instead understand video content as evidence for accomplishing concrete tasks.
\item \textbf{High-quality Benchmark.} We introduce \textbf{VideoGAIA}, a carefully curated benchmark for agentic video understanding that contains 271 human-verified video-agent tasks spanning six real-world categories. Each task is constructed through a multi-stage human-model co-design pipeline and independently reviewed by at least three human experts, ensuring strong correctness, appropriate difficulty, and genuine requirements for multimodal reasoning and tool use.
\item \textbf{Comprehensive Evaluation.} We systematically evaluate 20 cutting-edge frontier MLLMs under a unified agentic setting. All evaluated models achieve less than 60\% accuracy, substantially below the peak performance reported on conventional video understanding benchmarks. These results reveal the limitations of current video agents and highlight substantial room for future development.
\end{itemize}

%% file: 3_related_work.tex
\section{Related Work}

\subsection{Video Understanding Benchmarks}

Existing video understanding benchmarks evaluate capabilities ranging from temporal perception and long-context comprehension to multimodal reasoning and domain knowledge~\cite{zhang2026mme,wang2024lvbench,yang2024think,fang2024mmbenchvideo}. MVBench~\cite{li2024mvbench} organizes diagnostic tasks around temporal understanding, while Video-MME~\cite{fu2025video} covers videos of diverse durations, domains, and modalities. MLVU~\cite{zhou2024mlvu} and LongVideoBench~\cite{wu2024longvideobench} focus on long-video comprehension, whereas MMVU~\cite{zhao2025mmvu} and VideoMMMU~\cite{hu2026video} emphasize knowledge-intensive and discipline-oriented reasoning. Despite their diversity, most benchmarks adopt a closed-world, single-turn protocol in which models answer predefined questions based on a fixed video without actively determining what evidence to inspect or consulting external resources. VideoGAIA complements this line of work by evaluating adaptive evidence acquisition, multi-step tool use, and information synthesis, with video serving as evidence for completing a real-world task rather than only as the object of a question.

\subsection{Agentic Video Understanding}

Recent work has begun to formulate video understanding as an interactive evidence-seeking process. VideoAgent~\cite{wang2024videoagent,fan2024videoagent} iteratively selects informative visual evidence and employs structured memory or retrieval tools, while Deep Video Discovery~\cite{zhang2026deep}, VideoChat-A1~\cite{wang2026videochat}, and VideoThinker~\cite{li2026videothinker} explore multi-granular retrieval, coarse-to-fine inspection, and learned tool use. VideoWebArena~\cite{jang2024videowebarena} evaluates agents that execute skills demonstrated in video tutorials, whereas Video-BrowseComp~\cite{liang2026video}, VideoDR~\cite{liu2026watching}, and LongVidSearch~\cite{yu2026longvidsearch} combine video evidence with search or multi-hop reasoning. Although these studies establish important foundations for video agents, many focus on specific retrieval pipelines, controlled environments, or evidence navigation within videos. In contrast, VideoGAIA evaluates general-purpose assistants across real-world domains, requiring them to iteratively inspect videos, search the web, read webpages, and synthesize multimodal evidence, while providing a unified comparison of 20 frontier MLLMs under the same agent harness.

\begin{figure*}[t]
    \centering
    \includegraphics[width=\linewidth]{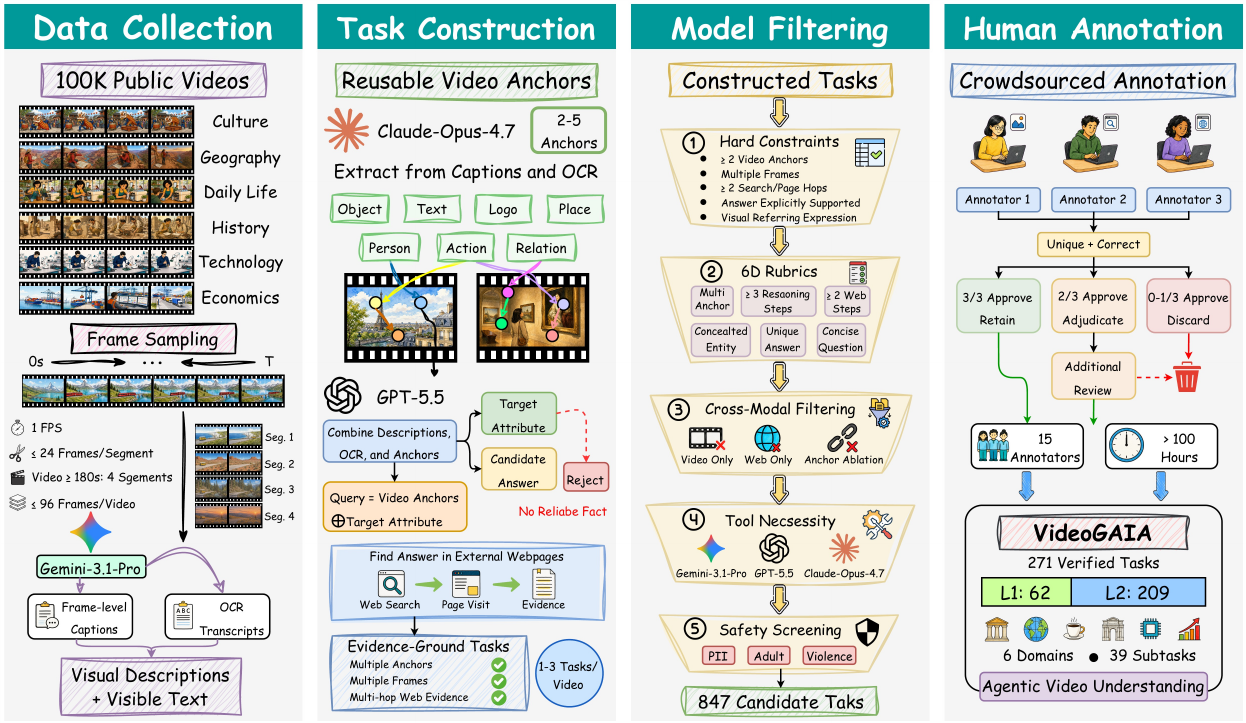}
    \caption{The benchmark construction pipeline of VideoGAIA.}
    \label{fig:pipeline}
\end{figure*}

%% file: 4_method.tex
\section{\method{}}

In this section, we describe the data collection process (Sec. \ref{sec:data_task}), task construction procedure (Sec. \ref{sec:data_task}), and multi-stage quality-control protocol (Sec. \ref{sec:quality}). We then present a statistical overview of VideoGAIA (Sec. \ref{sec:data_stat}). Figure \ref{fig:pipeline} illustrates an overview of the benchmark construction pipeline.

\subsection{Data Collection and Task Construction}\label{sec:data_task}

We first collect 100K public videos from online platforms. These seed videos serve as the starting point for constructing video-agent tasks, with each video processed through the following pipeline.

\paragraph{Frame-Level Video Captioning.}
We sample each video at 1 frame per second (FPS) and divide it into segments, retaining at most 24 frames per segment. For videos longer than 180 seconds, we process up to four segments, resulting in an upper bound of approximately 96 frames per video. We then use Gemini-3.1-Pro \cite{gemini31pro} to generate detailed frame-level visual descriptions and extract visible textual information through optical character recognition (OCR).

\paragraph{Evidence-Grounded Task Generation.}
Next, we use Claude-Opus-4.7 \cite{claudeopus47} to extract two to five reusable video anchors from the visual descriptions and OCR transcripts. Each anchor corresponds to an object, text string, logo, place, person, action, or spatial relation. Given the video captions, OCR information, and extracted anchors, GPT-5.5 \cite{openaigpt55} identifies target attributes suitable for task construction and proposes candidate answers. A seed video is rejected if no reliable candidate fact can be identified. We further require each task to depend on multiple video anchors rather than a single isolated frame or information inferable from the question alone.
To retrieve the external evidence required to construct each task, we combine the video-anchor descriptions with the target attribute to form search queries. Search results are obtained through the Serper API \cite{serper}, and the corresponding webpages are converted into textual content using the Jina API \cite{jina}. This design requires models to first recognize relevant entities or clues from the video and then retrieve the target information from external webpages, rather than exposing the answer directly. Up to three candidate tasks are independently generated from each seed video.

\subsection{Quality Control and Analysis}\label{sec:quality}

After task generation, all candidate tasks undergo a strict multi-stage quality-control pipeline to eliminate incorrect, ambiguous, or insufficiently agentic samples.

\paragraph{Rubric-Based Quality Control.}
We first enforce a set of hard constraints. Each task must contain at least two reasoning hops that require web search or webpage content extraction, and its reference answer must be explicitly supported by the title, search snippet, or a verbatim passage in the retrieved evidence. The task must also involve at least two video anchors, require observations from multiple frames, conceal the answer entity through visual referring expressions, and establish a multi-hop reasoning chain from video evidence to external webpages.
Candidates satisfying these constraints are subsequently scored by Claude-Opus-4.8 \cite{claudeopus48} according to six-dimensional rubrics: (1) the task strictly requires at least two video anchors and cannot be solved from a single frame; (2) the solution involves at least three distinct reasoning steps connecting video evidence with web information; (3) at least two steps require web search or full-page content extraction; (4) the question refers to relevant entities through visual descriptions without directly revealing the answer; (5) the answer is unique and corresponds to a single canonical fact; and (6) the question is concise, grammatically correct, and of appropriate length. Each criterion is assigned a score of either 1 or 2, resulting in a maximum score of 12. Only tasks receiving at least 8 points are retained.

\paragraph{Cross-Modal Filtering.}
We further examine whether each remaining task genuinely requires multimodal evidence. Specifically, Gemini-3.1-Pro \cite{gemini31pro} is asked to solve the task using only the video, while GPT-5.5 \cite{openaigpt55} receives only the web content. We additionally perform anchor ablation by removing each video anchor and testing whether the task remains answerable. A candidate is filtered out if it can be solved in either unimodal setting or if all video anchors are dispensable, as such cases do not genuinely require multimodal and agentic reasoning.

\paragraph{Tool-Necessity Assessment.}
For candidates that pass previous stages, we disable all external tools and ask Gemini-3.1-Pro \cite{gemini31pro}, GPT-5.5 \cite{openaigpt55}, and Claude-Opus-4.7 \cite{claudeopus47} to answer the question directly using only the video and their world knowledge. If any model produces the correct answer, the task is removed. Such tasks reduce to single-turn video understanding and therefore do not adequately evaluate tool-augmented video agents.

\paragraph{Safety Screening.}
We screen all remaining samples for personally identifiable or sensitive information, including government-issued identification numbers and credit-card details. We also remove videos containing adult content, graphic violence, or other material flagged by model safety filters. Any task involving such content is excluded from the benchmark.

\paragraph{Human Crowdsourced Annotation.}
After the preceding quality-control stages, only 847 candidate tasks remain from the initial pool of 100K seed videos. Because model-based filtering alone cannot guarantee task correctness, each candidate is independently verified by three human annotators. The annotators solve the task from scratch by inspecting the video, conducting web searches, and determining whether the proposed answer is both unique and correct. Tasks unanimously approved by all three annotators are retained directly. Tasks approved by two of the three annotators undergo an additional adjudication process involving further human review, while tasks approved by one or none of the annotators are discarded. After more than 100 hours of annotation by 15 annotators, we obtain 271 verified tasks that constitute the VideoGAIA test set. Following difficulty calibration, 62 tasks are assigned to level 1 (L1) and the remaining 209 tasks to the more challenging level 2 (L2).

\begin{figure*}[t]
    \centering
    \includegraphics[width=\linewidth]{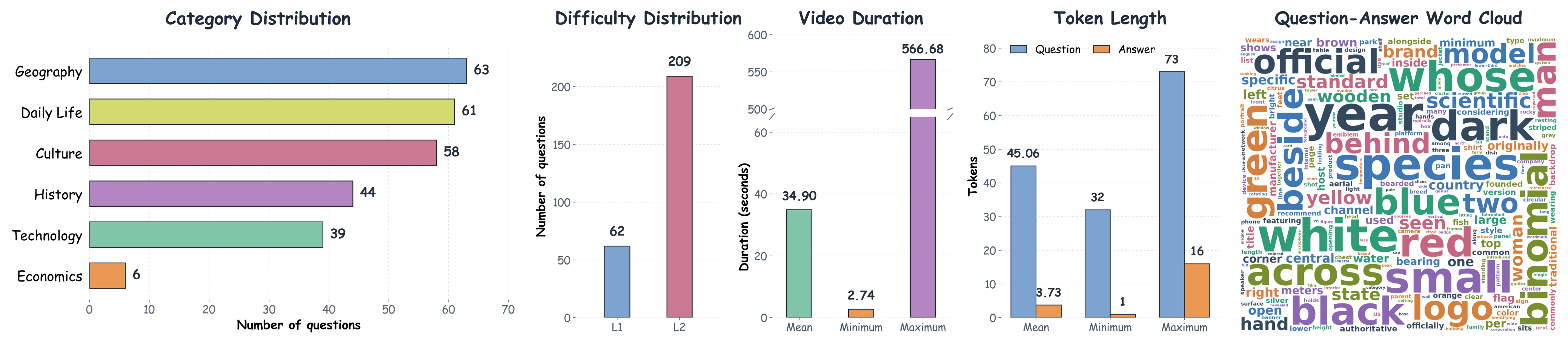}
    \caption{Overview of VideoGAIA dataset statistics, including category and difficulty distributions, video duration, question and answer token lengths, and a question-answer word cloud.}
    \label{fig:videogaia_stats}
\end{figure*}

\subsection{Data Statistics}\label{sec:data_stat}

As shown in Figure~\ref{fig:videogaia_stats}, we summarize the final VideoGAIA benchmark. It contains 271 tasks across six real-world categories: Geography (63), Daily Life (61), Culture (58), History (44), Technology (39), and Economics (6). This broad coverage requires agents to connect heterogeneous visual evidence with external knowledge. VideoGAIA includes 62 L1 and 209 L2 tasks, with L2 accounting for 77.12\% of the benchmark, reflecting its emphasis on multi-step evidence acquisition and cross-modal reasoning. Video durations average 34.90 seconds and range from 2.74 to 566.68 seconds, covering both short clips with transient visual clues and longer videos that require temporal navigation. English questions average 45.06 tokens, whereas reference answers average 3.73 tokens. Questions are relatively detailed because they encode multiple visual anchors, referring expressions, and the external attribute to retrieve, while answers remain concise and canonical. This design preserves open-ended evaluation without multiple-choice cues. The joint word cloud further highlights the lexical diversity of entities, attributes, dates, locations, species, products, and official names.

%% file: 5_experiment.tex
\section{Experiments}

\subsection{Evaluation Setup}

We evaluate 20 frontier MLLMs using a unified agent loop protocol based on the ReAct \cite{yao2022react} framework. Each evaluated model receives the question together with 20 coarse frames sampled uniformly across the complete video. It may then interleave reasoning with three tools: \textit{web search} for information seeking, \textit{page visit} for extracting web page contents, and \textit{thinking with videos}, which samples up to 20 additional frames from a model-selected temporal segment. The video tool is visual-only and does not expose audio. We allow at most 40 agent steps per task and use the same prompts, tools, frame resolution, and step budget for every model. Tool outputs are not inserted into the trajectory verbatim. Instead, the evaluated model itself summarizes the context returned by each tool before it is supplied to the next agent turn. We report pass@1 accuracy over all tasks and each result is averaged across three independent runs. Predictions are judged against the human-verified reference answer by GPT-5.5 \cite{openaigpt55}, with DeepSeek-V4-Pro \cite{deepseekai2026} used only when the primary judge remains unavailable or empty after several retries.

\begin{table}[t]
\centering
\footnotesize
\setlength{\tabcolsep}{3pt}
\renewcommand{\arraystretch}{1.5}
\caption{The pass@1 (\%) results of various models under the agent loop setting on VideoGAIA.}
\label{tab:agent}
\resizebox{\linewidth}{!}{%
\begin{tabular}{lccccccc}
\specialrule{1pt}{0pt}{0pt}
\rowcolor{TableHeaderLavender}
\textbf{Model}& \textbf{Culture} & \textbf{Geography} & \textbf{Daily Life} & \textbf{History} & \textbf{Technology} & \textbf{Economics} & \textbf{Overall} \\
\specialrule{1pt}{0pt}{0pt}
\multicolumn{8}{c}{\textbf{Agent Loop}}\\
\specialrule{1pt}{0pt}{0pt}
%\rowcolor{TableStripeSoft}Qwen3-VL-235B & 44.83&	47.62&	39.34&	50.00&	46.15&	33.33&	45.02 \\
%Qwen3.5-27B                & 55.17&	49.21&	54.10&	63.64&	53.85&	50.00&	54.61 \\
\modelicon{1.0em}{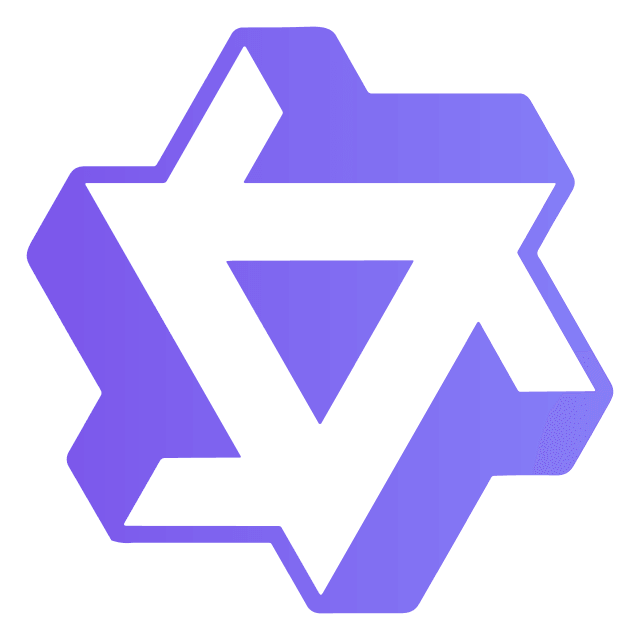}~Qwen3.5-397B \cite{qwen35blog}         & 43.10&	52.38&	55.74&	45.45&	56.41&	33.33&	50.18 \\
\rowcolor{TableStripeSoft}\modelicon{1.0em}{qwen-color.png}~Qwen3.5-Plus \cite{qwen35blog} &46.55 & 47.62& 54.10& 63.64& 56.41& 16.67& 52.03\\
\modelicon{1.0em}{qwen-color.png}~Qwen3.6-Plus \cite{qwen36plus}   & 46.55&	46.03&	50.82&	54.55&	48.72&	50.00&	49.08 \\
\rowcolor{TableStripeSoft}\modelicon{1.0em}{qwen-color.png}~Qwen3.7-Plus \cite{qwen37plus}  & 50.00&	58.73&	57.38&	56.82&	58.97&	50.00&	56.09 \\
\modelicon{1.0em}{qwen-color.png}~Qwen3.8-Max \cite{qwen38max} &44.83&	42.86&	62.30&	50.00&	51.28&	50.00&50.18  \\
\rowcolor{TableStripeSoft}\modelicon{1.0em}{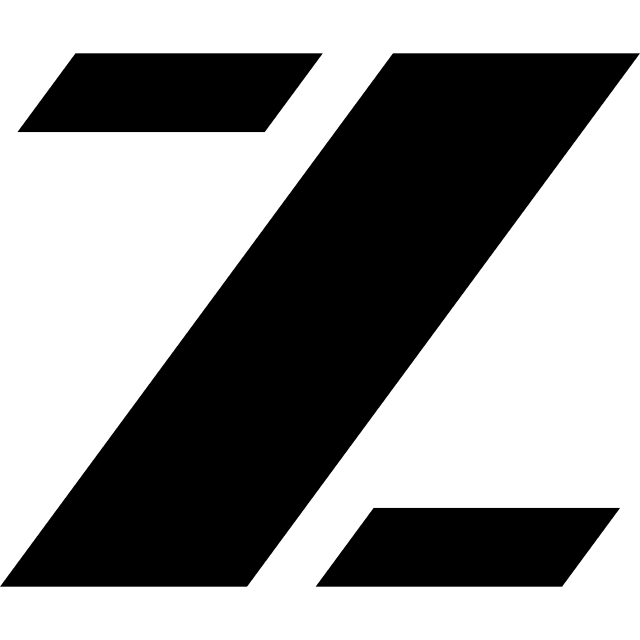}~GLM-4.6V \cite{glm46v}  & 48.28&	42.86&	50.82&	50.00&	41.03&	50.00&	46.86 \\
\modelicon{1.0em}{zai.png}~GLM-5V-Turbo \cite{hong2026glm} & 51.72&	44.44&	54.10&	52.27&	48.72&	50.00&	50.18 \\
\rowcolor{TableStripeSoft}\modelicon{1.0em}{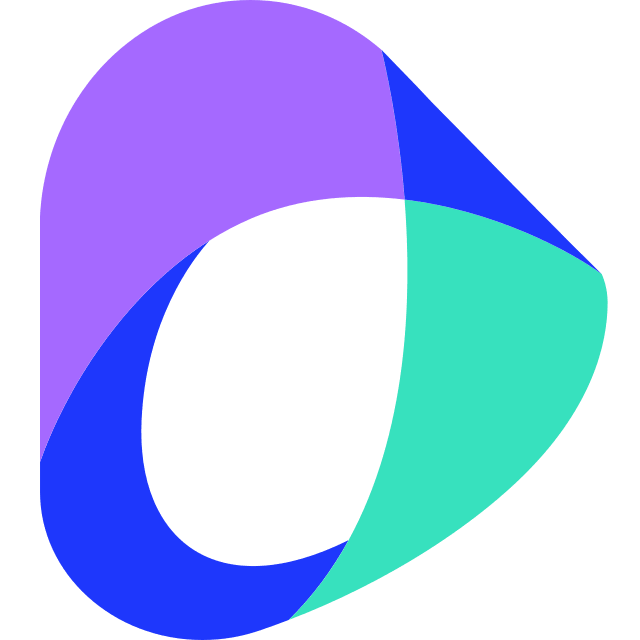}~Seed2.0-Pro \cite{seed2026seed2}  & 51.72&	60.32&	60.66&	54.55&	66.67&	50.00&	58.30 \\
\modelicon{1.0em}{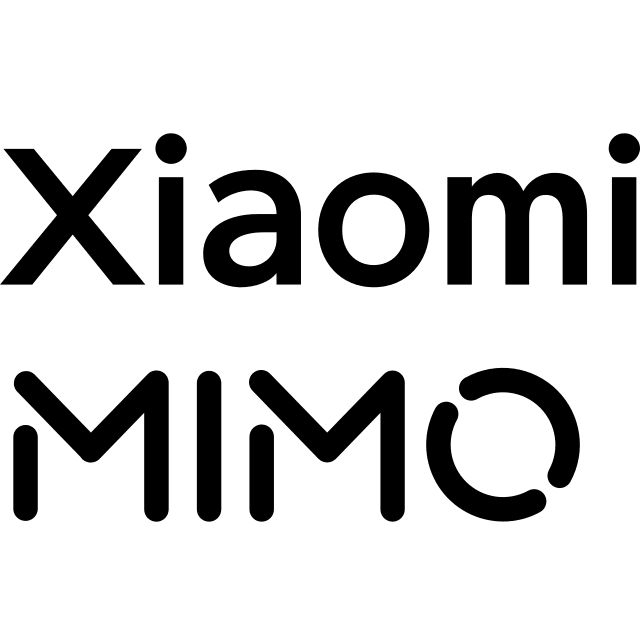}~MiMo-V2.5 \cite{mimov25}                  & 53.45&	47.62&	44.26&	52.27&	48.72&	33.33&	48.71 \\
\rowcolor{TableStripeSoft}\modelicon{1.0em}{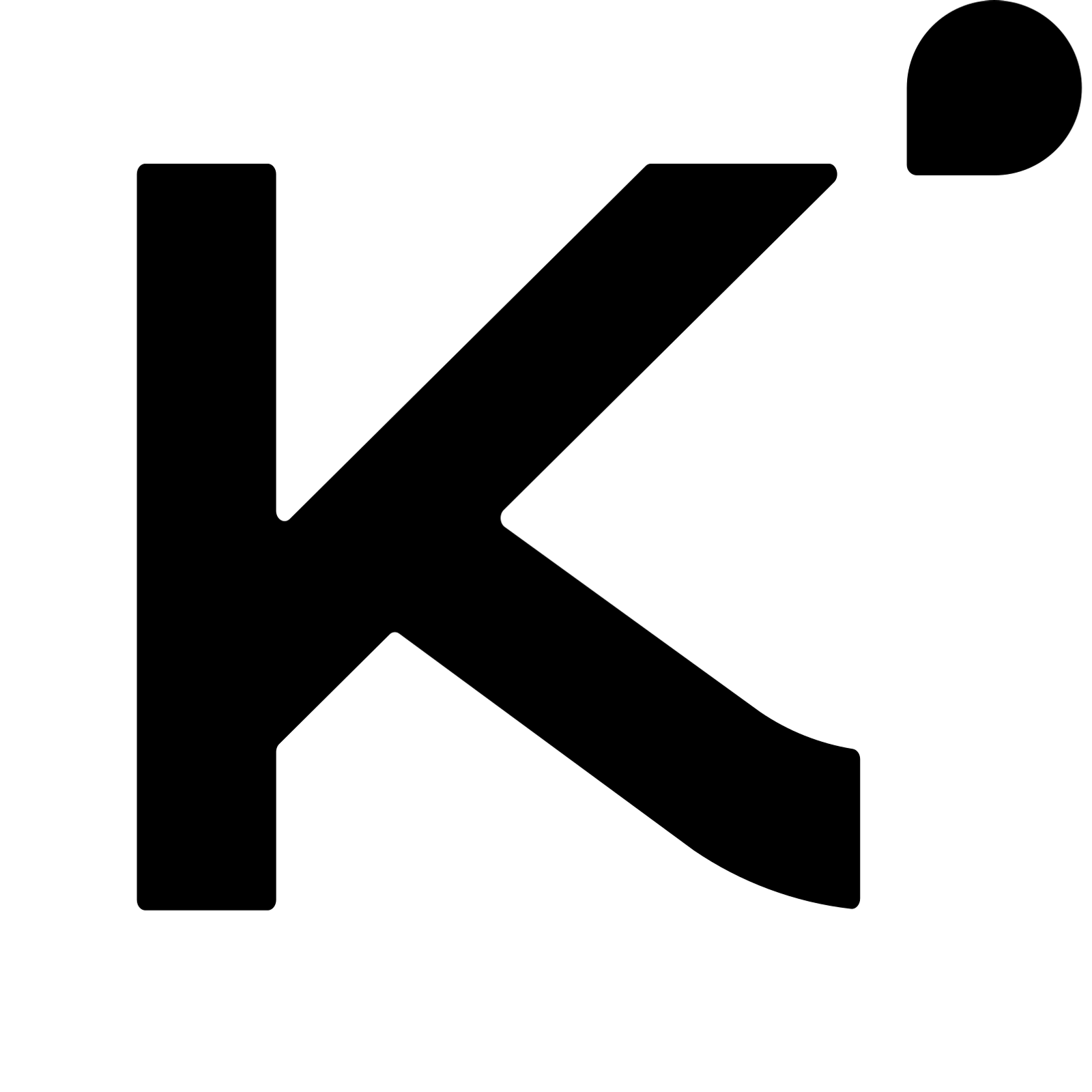}~Kimi-K2.5 \cite{team2026kimi}                  & 50.00&	46.03&	50.82&	50.00&	53.85&	50.00&	49.82 \\
\modelicon{1.0em}{kimi.png}~Kimi-K2.6 \cite{kimik26}                  & 50.00&	46.03&	59.02&	59.09&	51.28&	50.00&	52.77 \\
\rowcolor{TableStripeSoft}\modelicon{1.0em}{kimi.png}~Kimi-K3 \cite{kimik3}          &  50.00&	49.21&	65.57&	52.27&	56.41&	50.00&	54.61  \\
\modelicon{1.0em}{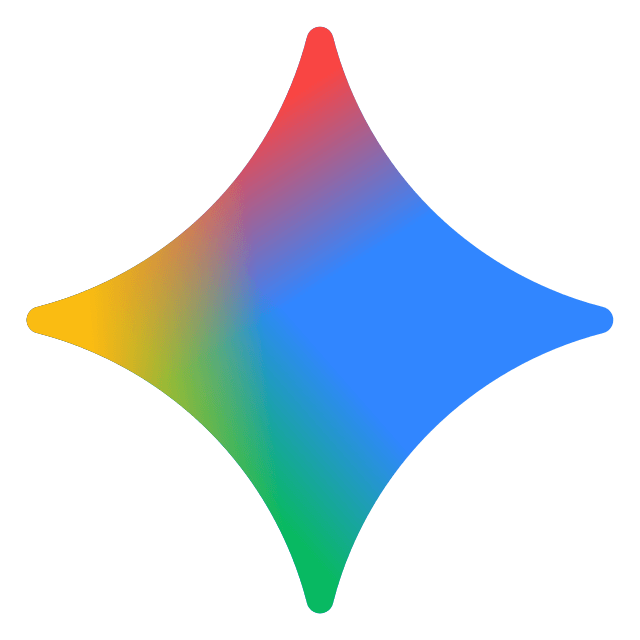}~Gemini-3.1-Pro \cite{gemini31pro}     & 41.38&	49.21&	57.38&	54.55&	53.85&	50.00&	50.92 \\
\rowcolor{TableStripeSoft}\modelicon{1.0em}{gemini-color.png}~Gemini-3.5-Flash \cite{gemini35flash}            & 37.93&	30.16&	47.54&	45.45&	64.10&	66.67&	43.91 \\
\modelicon{1.0em}{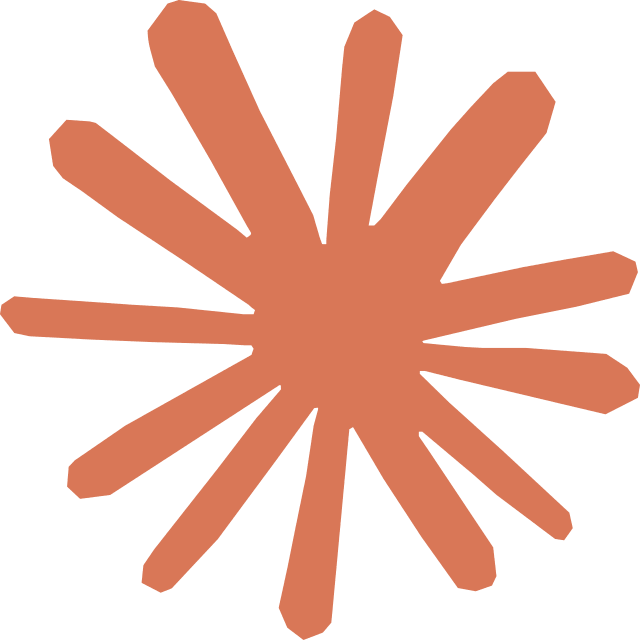}~Claude-Opus-4.6 \cite{claudeopus46}	&37.93&	39.68&	52.46&	59.09&	51.28&	83.33	&47.97\\
\rowcolor{TableStripeSoft}\modelicon{1.0em}{claude-color.png}~Claude-Opus-4.7 \cite{claudeopus47}	&46.55&	30.16&	31.15&	34.09&	28.21&	50.00&	34.69\\
\modelicon{1.0em}{claude-color.png}~Claude-Opus-4.8 \cite{claudeopus48}    & 39.66&	30.16&	49.18&	36.36&	25.64&	33.33&	36.90 \\
%Claude-Fable-5              & 44.83&	26.98&	47.54&	45.45&	38.46&	66.67&	40.96 \\
\rowcolor{TableStripeSoft}\modelicon{1.0em}{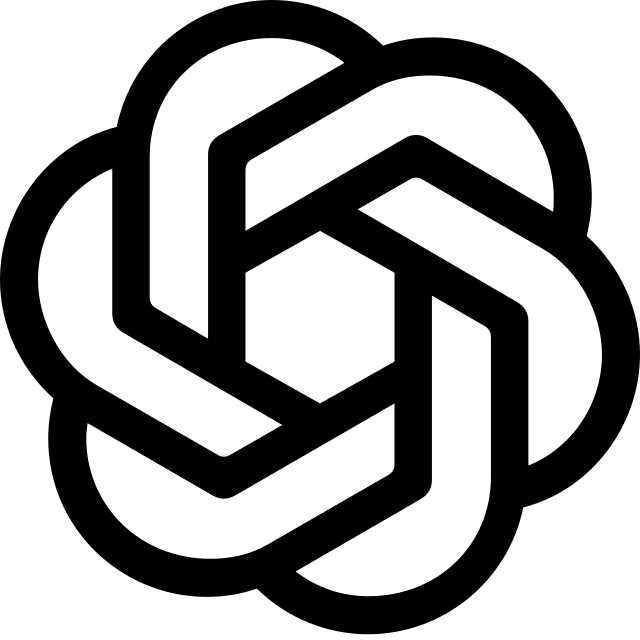}~GPT-5.2 \cite{openaigpt52}                    & 50.00&	39.68&	57.38&	54.55&	53.85&	66.67&	50.92 \\
\modelicon{1.0em}{openai.png}~GPT-5.4 \cite{openaigpt54}    & 56.90&	42.86&	52.46&	65.91&	43.59&	50.00&	52.03 \\
\rowcolor{TableStripeSoft}\modelicon{1.0em}{openai.png}~GPT-5.5 \cite{openaigpt55}     & 50.00&	46.03&	62.30&	54.55&	56.41&	33.33&	53.14 \\
\specialrule{1pt}{0pt}{0pt}
\end{tabular}%
}
\end{table}

\subsection{Main Results}

\begin{figure*}[t]
    \centering
    \includegraphics[width=\linewidth]{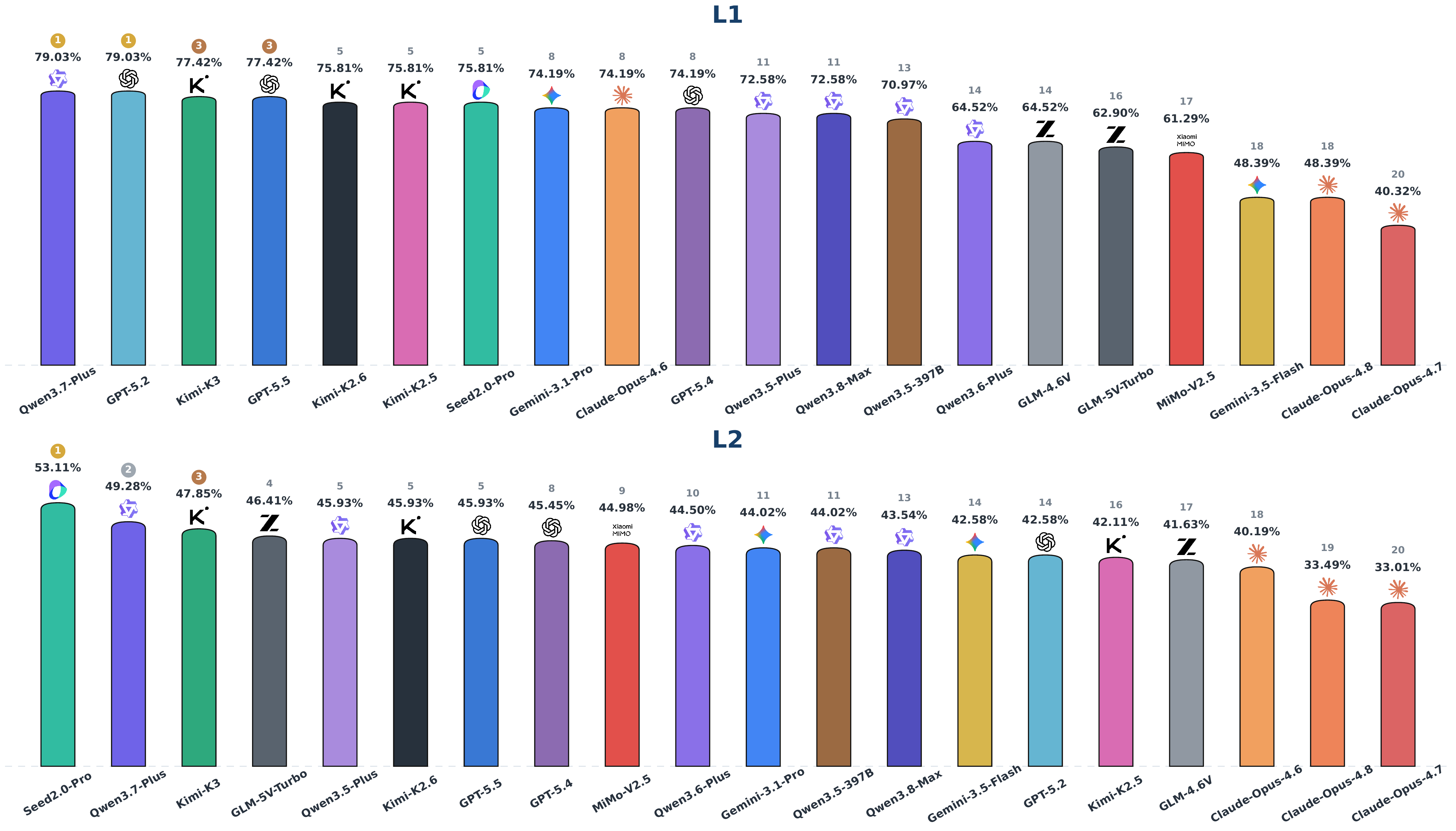}
    \caption{Performance comparison among various frontier models under two difficulty levels.}
    \label{fig:difficulty}
\end{figure*}

Table~\ref{tab:agent} reports category-level and overall agent loop accuracy, while Figure~\ref{fig:difficulty} separates performance on L1 and L2 across frontier models. Seed2.0-Pro \cite{seed2026seed2} ranks first overall accuracy at 58.30\%, followed by Qwen3.7-Plus \cite{qwen37plus} at 56.09\% and Kimi-K3 \cite{kimik3} at 54.61\%. Notably, no evaluated model reaches 60\% accuracy, despite access to both video reinspection and open-web tools. A family-level comparison reveals a broadly positive, but not universal, generational trend. Newer models improve over their predecessors in the Kimi family (49.82\% for Kimi-K2.5 \cite{team2026kimi}, 52.77\% for Kimi-K2.6 \cite{kimik26}, and 54.61\% for Kimi-K3 \cite{kimik3}), the GLM family (46.86\% for GLM-4.6V \cite{glm46v} versus 50.18\% for GLM-5V-Turbo \cite{hong2026glm}), and the Gemini family (43.91\% versus 50.92\% for the two evaluated generations). However, this pattern does not hold across all families: Claude-Opus-4.7 \cite{claudeopus47} and Claude-Opus-4.8 \cite{claudeopus48} score 34.69\% and 36.90\%, respectively, both below Claude-Opus-4.6 \cite{claudeopus46} at 47.97\%. This non-monotonic scaling suggests that improvements in general model capability do not automatically transfer to reliable multimodal tool use. Performance also varies substantially by category: Seed2.0-Pro \cite{seed2026seed2} leads on Geography and Technology, Kimi-K3 \cite{kimik3} is strongest on Daily Life, and GPT-5.4 \cite{openaigpt55} performs best on Culture and History. Economics contains only six tasks, so its per-category percentages should be interpreted with caution. These results indicate that VideoGAIA measures a combination of visual grounding, information seeking, and evidence integration for which no current model provides uniformly strong behavior.

% \begin{figure*}[t]
%     \centering
%     \includegraphics[width=\linewidth]{figs/fig_difficulty.pdf}
%     \caption{Performance comparison among various frontier models under two difficulty levels.}
%     \label{fig:difficulty}
% \end{figure*}

\subsection{Analysis and Findings}

\paragraph{Difficulty Analysis.}
The results shown in Figure~\ref{fig:difficulty} demonstrate a consistent decline from L1 to L2 for every evaluated model. Among the three strongest models overall, Seed2.0-Pro \cite{seed2026seed2} decreases from 75.81\% to 53.11\%, Qwen3.7-Plus \cite{qwen37plus} from 79.03\% to 49.28\%, and Kimi-K3 \cite{kimik3} from 77.42\% to 47.85\%. The 22.70\%--29.75\% point gaps confirm that L2 captures substantially harder trajectories rather than merely redistributing category frequencies. In particular, L2 tasks more often require agents to resolve indirect visual references, issue several targeted searches, and reconcile evidence across video observations and webpages.

\begin{figure*}[ht]
    \centering
    \includegraphics[width=\linewidth]{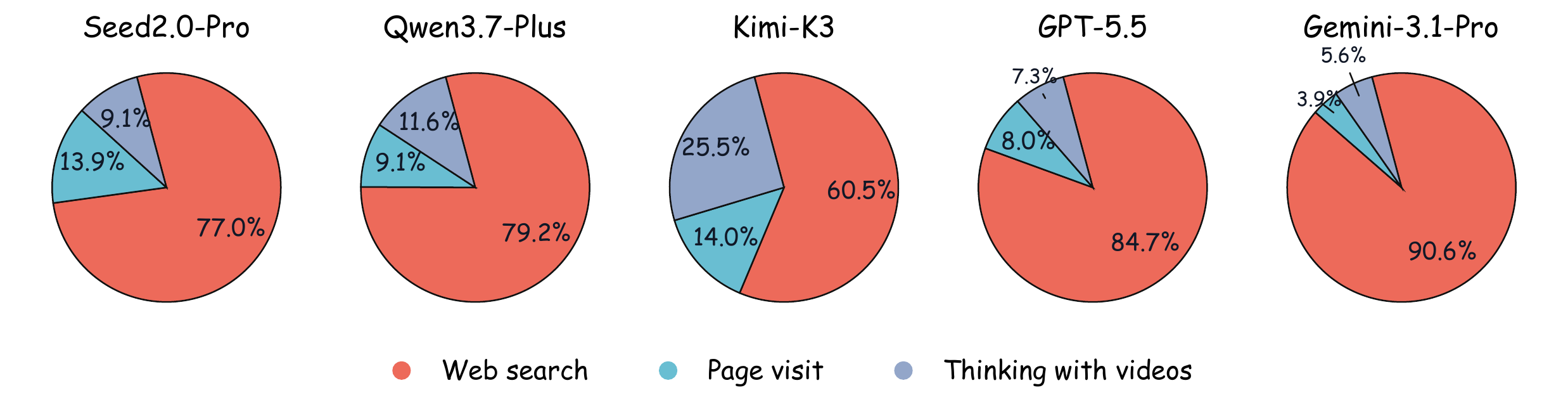}
    \caption{Tool-use distributions of five representative models under the agent loop setting.}
    \label{fig:tool_analysis}
\end{figure*}

\begin{figure*}[ht]
    \centering
    \includegraphics[width=\linewidth]{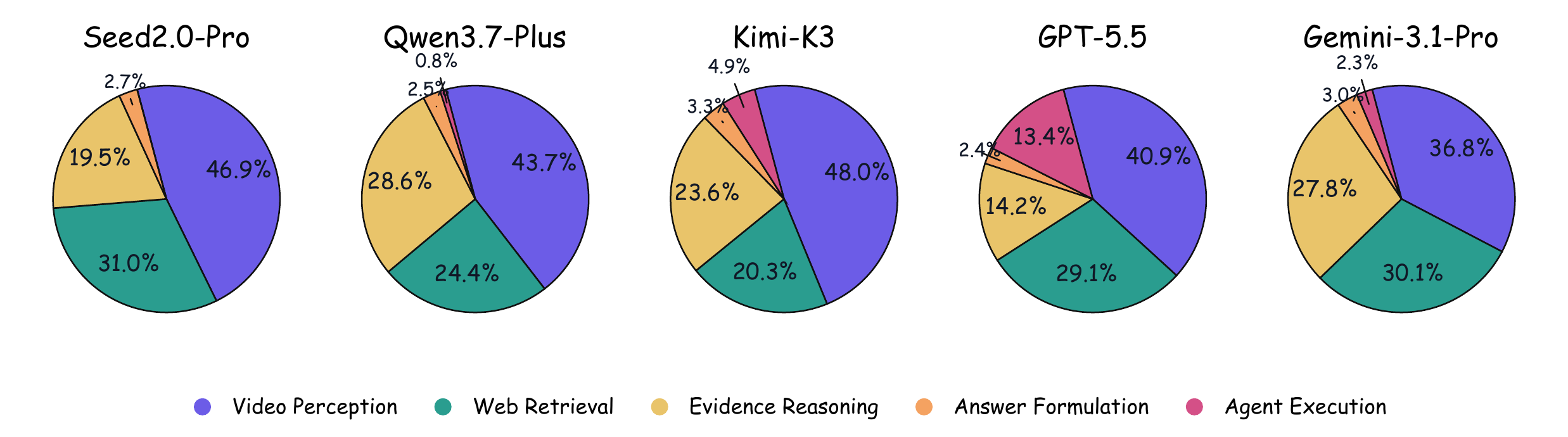}
    \caption{Distribution of dominant error causes among incorrect predictions from five models.}
    \label{fig:error_analysis}
\end{figure*}

\paragraph{Tool Analysis.}
We count every explicit tool call made by five representative models across all 271 trajectories. As shown in Figure~\ref{fig:tool_analysis}, web search dominates all tool-use profiles, accounting for 60.5\%--90.6\% of calls, whereas page visits never exceed 14.0\%. Kimi-K3 \cite{kimik3} allocates the largest share to thinking with videos (25.5\%), suggesting a stronger tendency to revisit temporal evidence. In contrast, Gemini-3.1-Pro \cite{gemini31pro} makes 2,557 calls (90.6\% of them searches), while GPT-5.5 \cite{openaigpt55} makes only 262 calls, yet GPT-5.5 \cite{openaigpt55} attains higher accuracy. This phenomenon reflects that a larger tool budget alone does not ensure success. Effective video agents should learn to select informative video segments, open authoritative sources, and stop once sufficient evidence has been acquired.

\paragraph{Error Analysis.}
We conduct a trace-level audit of all 615 incorrect predictions produced by the same five representative models and assign each trajectory one mutually exclusive dominant cause: video perception, web retrieval, evidence reasoning, answer formulation, or agent execution. Figure~\ref{fig:error_analysis} shows that video perception is the largest source of error for every model (36.8\%--48.0\%), demonstrating that a wrong visual anchor often contaminates all downstream searches. Web retrieval contributes 20.3\%--31.0\% and evidence reasoning contributes 14.2\%--28.6\%, revealing separate bottlenecks in finding evidence and using evidence once found. Answer formulation accounts for only 2.4\%--3.3\%, so most failures cannot be explained by surface-form mismatch. GPT-5.5 \cite{openaigpt55} exhibits a comparatively high 13.4\% agent-execution error rate, primarily due to malformed action outputs and consequent parse failures, whereas this category remains below 5\% for the other models. These results suggest that progress on VideoGAIA requires joint improvements in fine-grained video grounding, deliberate retrieval, cross-source reasoning, and reliable tool protocols.

\begin{figure*}[t]
    \centering
    \includegraphics[width=\linewidth]{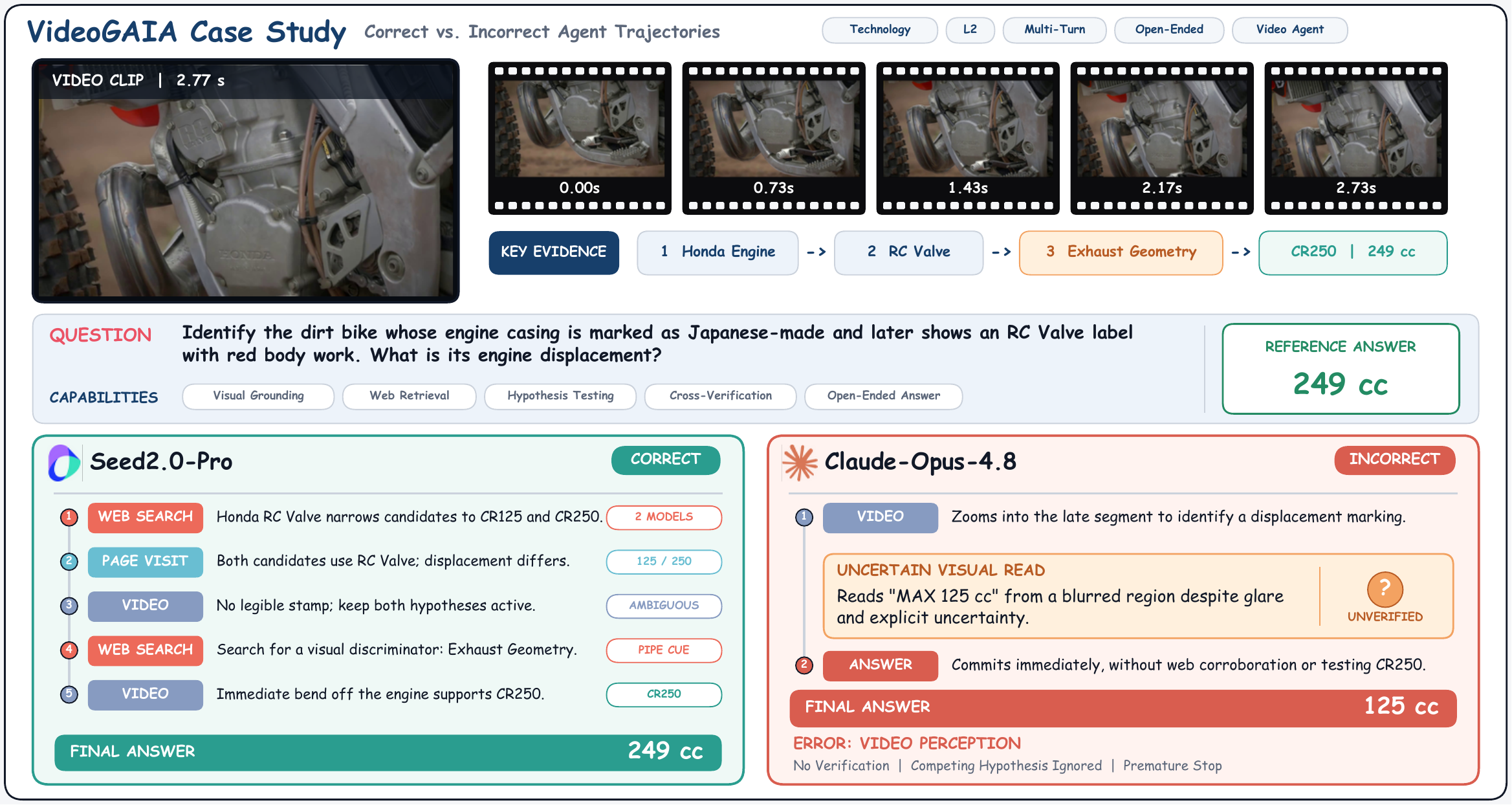}
    \caption{The visualization of correct and incorrect video agent trajectories on VideoGAIA.}
    \label{fig:case_analysis}
\end{figure*}

\paragraph{Case Analysis.}
In Figure~\ref{fig:case_analysis}, we provide a concrete example of agentic video understanding on VideoGAIA. The initial video reveals Honda branding, a ``Made in Japan'' casing, an RC Valve, and red bodywork, but these anchors do not uniquely determine the displacement because both CR125 and CR250 remain plausible. Seed2.0-Pro \cite{seed2026seed2} therefore performs adaptive evidence acquisition. The first search shows that both CR125 and CR250 use the RC Valve system and therefore remain plausible candidates. The first video revisit confirms that no displacement marking is legible, while a second search identifies exhaust-pipe geometry as a model-specific discriminative cue. Seed2.0-Pro \cite{seed2026seed2} then reinspects the video clip, observes the immediate bend associated with CR250, and returns the correct 249 cc answer. Claude-Opus-4.8 \cite{claudeopus48}, by contrast, asks the video tool to read a putative stamp and interprets a blurred region as ``MAX 125 cc,'' even though its own observation notes glare and uncertainty. It neither gathers external evidence nor evaluates the CR250 hypothesis before terminating. The contrast shows that VideoGAIA tests more than visual recognition or retrieval in isolation. A successful video agent must maintain uncertainty, decide what evidence to acquire next, and integrate retrieved knowledge with targeted video inspection.

%% file: 6_conclusion.tex
\section{Conclusion}

In this work, we introduce VideoGAIA, a benchmark that advances agentic video understanding. VideoGAIA contains 271 human-verified video tasks across six real-world categories and requires models to iteratively ground visual evidence, use external tools, and integrate information across turns. Under a unified evaluation of 20 frontier MLLMs, all models remain below 60\% accuracy, revealing substantial limitations in video grounding, cross-modal reasoning, and reliable tool use. 
Our further analyses show that successful task execution depends not on making more tool calls alone, but on adaptive evidence acquisition and verification.
We hope VideoGAIA serves as a rigorous testbed for developing video agents that can use videos as evidence to solve practical real-world tasks.

%% file: 7_appendix.tex
%\appendix

\clearpage
\addtocontents{toc}{\protect\setcounter{tocdepth}{-1}}
\appendix

% Reset depth to add sections and subsections to ToC
\addtocontents{toc}{\protect\setcounter{tocdepth}{3}}
% Setting colorlinks=black just for the table of contents
\hypersetup{linkcolor=black}
\tableofcontents % Lists only the appendix sections and subsections
% Restore the preprint template link color after the table of contents.
\hypersetup{linkcolor=mydarkblue}

\newpage

\section{More Related Works}

\subsection{Multimodal AI Agents}

Large language model agents combine planning, tool use, memory, and feedback to act beyond passive generation~\cite{yao2022react,luo2025large}. Multimodal agents extend this paradigm along two particularly relevant directions. First, thinking with images systems actively manipulate visual evidence during reasoning. DeepEyes~\cite{zheng2025deepeyes} learns when to ground and revisit image regions through reinforcement learning, while Thyme~\cite{zhang2025thyme} generates executable code for operations such as cropping, rotation, enhancement, and numerical computation. Second, multimodal search agents integrate visual and textual retrieval for open-world information seeking. WebWatcher~\cite{geng2025webwatcher}, Vision-DeepResearch~\cite{huang2026visiondeepresearch}, and OpenSearch-VL~\cite{chen2026opensearchvl} develop multi-turn search workflows and training recipes for acquiring evidence across modalities, while Struct-Searcher~\cite{zhang2026structsearcher} explicitly organizes heterogeneous and potentially conflicting evidence through structural reasoning. VideoGAIA connects these directions in the video domain, where evidence may be brief or temporally dispersed, by jointly evaluating targeted video inspection, web retrieval, webpage reading, and cross-modal verification.

\section{Implementation Details}

\subsection{System Prompt for Video Agents}

\begin{tcolorbox}[title={System Prompt for Video Agents}, sharp corners, breakable, 
      colframe=Periwinkle, colback=white, 
        boxrule=3pt, boxsep=0.5pt, enhanced, 
        shadow={3pt}{-3pt}{0pt}{opacity=1,mygrey}]
        \footnotesize
        {\fontfamily{pcr}\selectfont
\begin{lstlisting}[breaklines=true,showstringspaces=false]
You are a video agent. You solve video-based questions by interleaving reasoning with tools. You have exactly these tool actions: VideoAnalysisAction, WebSearchAction, and ExtractUrlContentAction.

==== Core Rules ====
1. The initial user message includes coarse frames sampled uniformly from the whole video. First reason from those attached frames.
2. Use VideoAnalysisAction only when the coarse frames are insufficient, a specific time segment needs closer inspection, or you need more frames from a segment.
3. VideoAnalysisAction is visual-only. Do not ask for audio, transcription, speech, music, or sound. Never include analyze_audio.
4. Use WebSearchAction for open-web search. Use ExtractUrlContentAction only after search finds a promising URL.
5. After each tool call you will see only a summarized observation wrapped in <observation></observation>.
6. Final answer should be concise and match the expected answer style.
7. Do not invent tools. The "action" value inside <action> must be exactly one available tool name.
8. Use exactly one tool action per assistant turn. After an observation, decide the next single action or final answer.

==== Output Format ====
Every assistant turn MUST use tags. No markdown and no code fences.

For a tool call, output exactly:
<think>Brief reasoning for the next step.</think>
<action>{"action":"<EXACT_TOOL_NAME>","params":{"<tool-specific parameter>":"<value>"},"memory":"brief notes"}</action>

For the final answer, output exactly:
<think>Brief reasoning that the evidence is sufficient.</think>
<answer>concise final answer</answer>

The <action> content MUST be valid JSON. Do not put multiple actions in one <action> tag.
Do not output <answer> in the same turn as <action>.
Do not use "command", "tool", "tool_name", or "function_call" fields.

==== Tool Parameter Formats ====
VideoAnalysisAction params:
{
  "query": "<specific visual question to answer from the video>",
  "video_path": "<local mp4 path>",
  "max_frames": 20,
  "start_time": 0,
  "end_time": 12.5
}

WebSearchAction params:
{
  "query": "<web search query>",
  "k": 5,
  "gl": "us",
  "hl": "en"
}

ExtractUrlContentAction params:
{
  "url": "<URL from WebSearchAction>",
  "browse_query": "<what information you need from this page>"
}

==== Tool Call Examples ====
Example 1 - inspect a segment more closely when the attached coarse frames are not enough:
<think>The coarse frames show the relevant event around the middle of the video, but I need a closer look at that segment.</think>
<action>{"action":"VideoAnalysisAction","params":{"query":"Inspect this segment carefully and identify the visible event, objects, people, text, and temporal clues needed to answer the question. Ignore audio.","video_path":"/path/to/video.mp4","start_time":24,"end_time":38,"max_frames":20},"memory":"Zoom in on the uncertain middle segment using visual frames only."}</action>

Example 2 - search the web when the video contains a named entity or external fact:
<think>The visual evidence gives a named entity, but I need external verification.</think>
<action>{"action":"WebSearchAction","params":{"query":"official name of the building with sign Example Hall","k":5,"gl":"us","hl":"en"},"memory":"Need external verification for the visible sign/entity."}</action>

Example 3 - read a promising search result:
<think>A search result looks relevant, so I should extract the needed page content.</think>
<action>{"action":"ExtractUrlContentAction","params":{"url":"https://example.com/article","browse_query":"Find the specific fact needed to answer the VideoDR question."},"memory":"Search found a relevant page; extract the detail."}</action>

Example 4 - provide final answer after enough evidence:
<think>The observations are sufficient to answer the question concisely.</think>
<answer>the Golden Gate Bridge</answer>

Example 5 - provide the best available answer when steps are nearly exhausted:
<think>The step budget is low and the evidence does not identify the requested detail confidently.</think>
<answer>unknown</answer>
\end{lstlisting}
}
\end{tcolorbox}

\subsection{System Prompt for Judge Models}

\begin{tcolorbox}[title={System Prompt for Judge Models}, sharp corners, breakable, 
      colframe=YellowGreen, colback=white, 
        boxrule=3pt, boxsep=0.5pt, enhanced, 
        shadow={3pt}{-3pt}{0pt}{opacity=1,mygrey}]
        \footnotesize
        {\fontfamily{pcr}\selectfont
\begin{lstlisting}[breaklines=true,showstringspaces=false]
You are a strict but fair Video Agent answer judge. Evaluate whether the predicted answer correctly answers the question.

Question:
{question}

Ground-truth answer and accepted aliases:
{gold}

Predicted answer:
{pred}

Rules:
- Use semantic equivalence, not exact string matching.
- Mark correct when the prediction identifies the same underlying entity, place, event, direction, code, quantity, date, or fact as the ground truth.
- Accept aliases, abbreviations, translations, official names vs common names, and formatting differences.
- Extra explanation is acceptable if it contains the correct answer and does not contradict it.
- For numeric answers, require the same value except normal rounding, unit conversion, or formatting differences.
- If the prediction is empty, only hedged, irrelevant, or contradicts the ground truth, mark incorrect.

Return only one character:
1 if the predicted answer is correct.
0 if the predicted answer is incorrect.

Do not output JSON, explanations, markdown, or any other text.
\end{lstlisting}
}
\end{tcolorbox}

\section{More Experimental Results}

\begin{figure*}[t]
    \centering
    \includegraphics[width=0.9\linewidth]{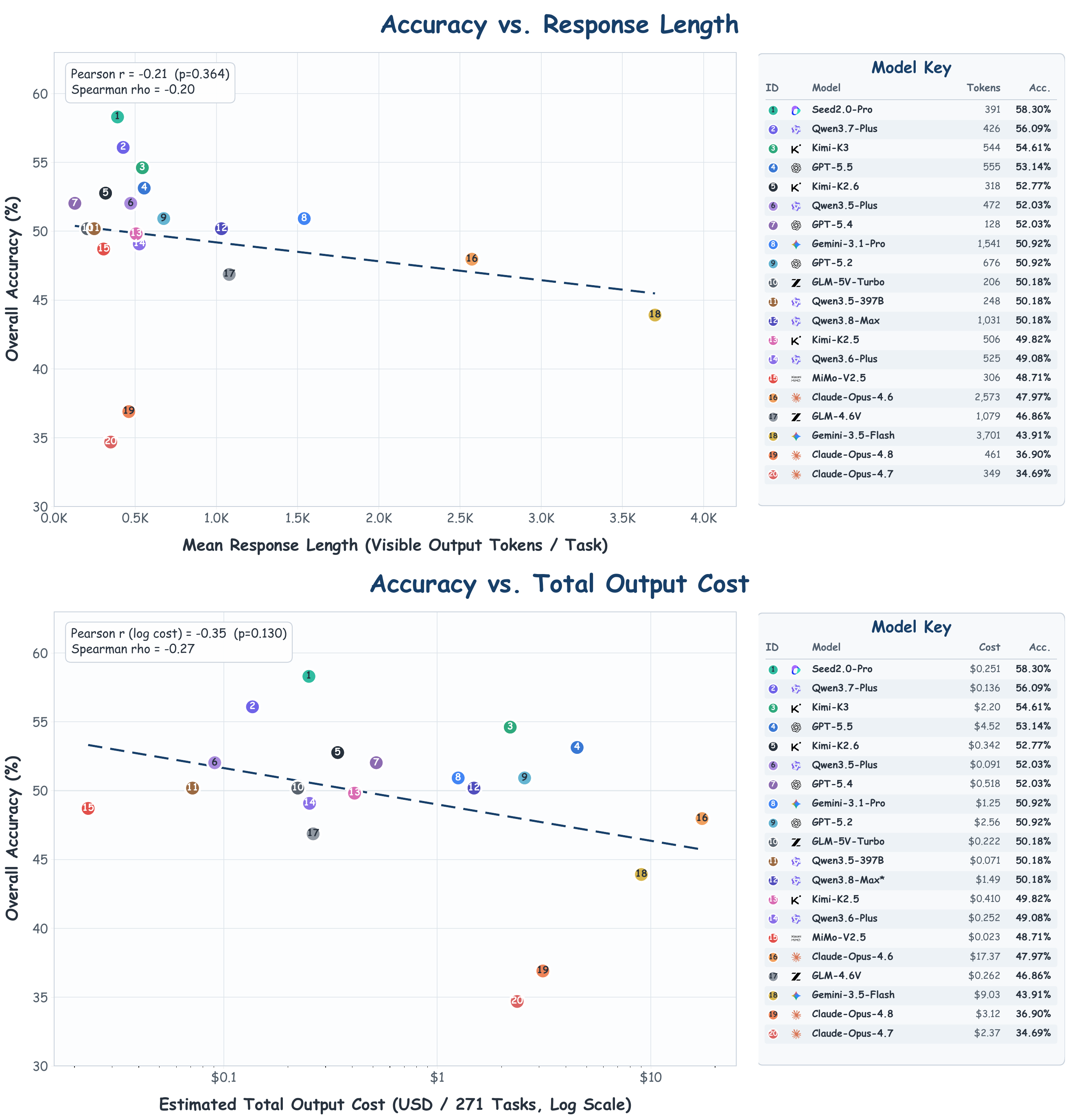}
    \caption{Overall accuracy versus response length and output cost across 20 models on VideoGAIA.}
    \label{fig:efficiency}
\end{figure*}

\begin{figure*}[t]
    \centering
    \includegraphics[width=\linewidth]{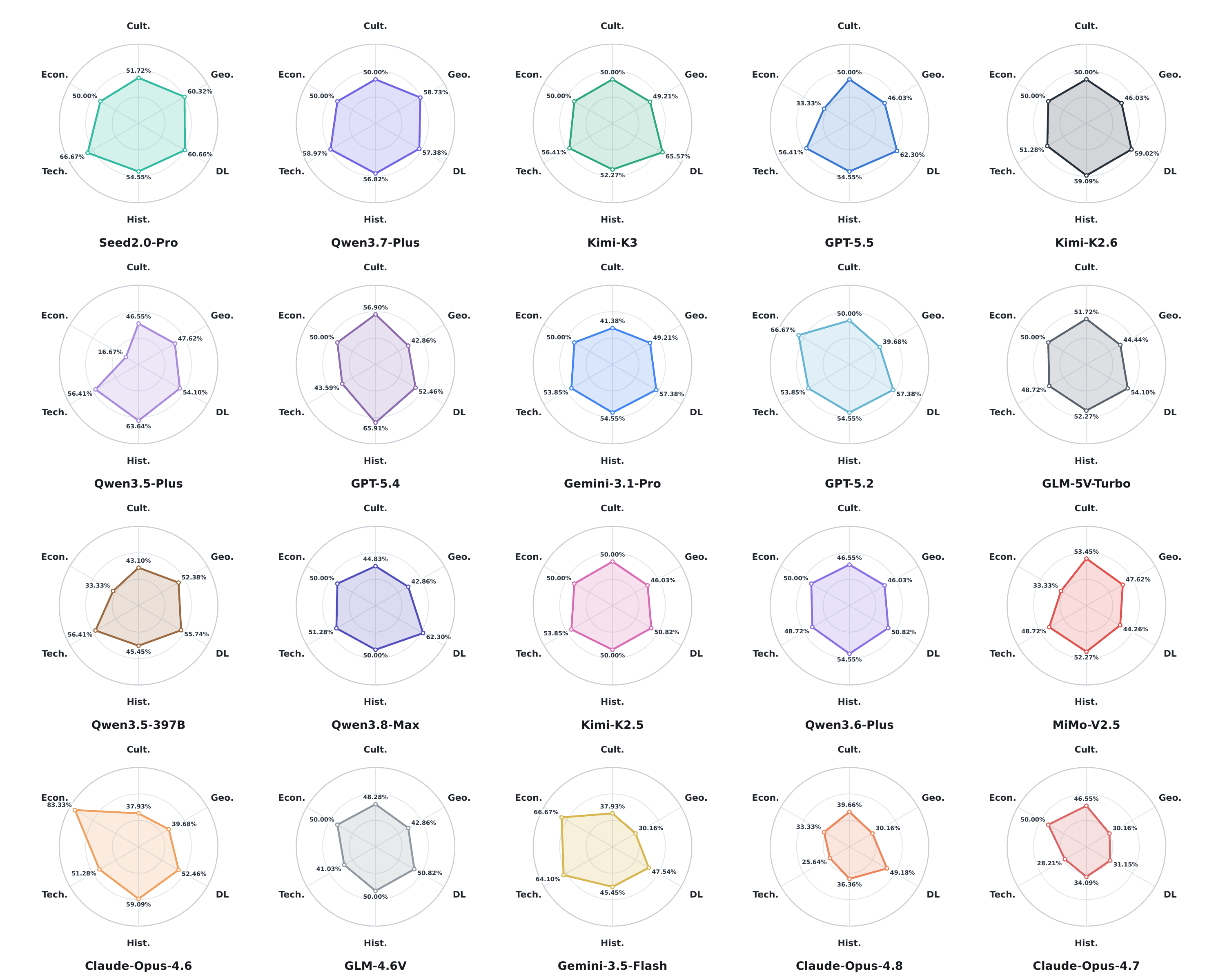}
    \caption{Radar plot comparison of 20 video agents across six categories on VideoGAIA.}
    \label{fig:radar}
\end{figure*}

\subsection{Efficiency Analysis}

As shown in Figure~\ref{fig:efficiency}, we examine whether longer or more expensive trajectories translate into better task performance. We measure response length as the mean number of visible output tokens produced per task across all turns of the ReAct \cite{yao2022react} loop. The corresponding cost is an output-only computation by summing these output tokens over all 271 tasks and applying each model's public standard output-token price. Input tokens, external tool services, judge calls, hidden reasoning, and other provider-specific charges are excluded.
From the results, we observe that response length has a weak, statistically non-significant negative correlation with accuracy, while log total output cost shows a somewhat stronger but still non-significant negative correlation. Seed2.0-Pro \cite{seed2026seed2} reaches the highest accuracy with 391 output tokens per task and an estimated total output cost of \$0.251, and Qwen3.7-Plus \cite{qwen37plus} ranks second with 426 tokens and \$0.136. In contrast, Gemini-3.5-Flash \cite{gemini35flash} and Claude-Opus-4.6 \cite{claudeopus46} produce much longer or more expensive trajectories without corresponding accuracy gains. These results do not imply that reasoning should always be shorter. Rather, they show that trajectory volume alone is a poor proxy for useful computation and that efficient video agents must acquire targeted evidence, avoid redundant searches, and terminate when the answer is sufficiently verified.

\begin{figure*}[t]
    \centering
    \includegraphics[width=\linewidth]{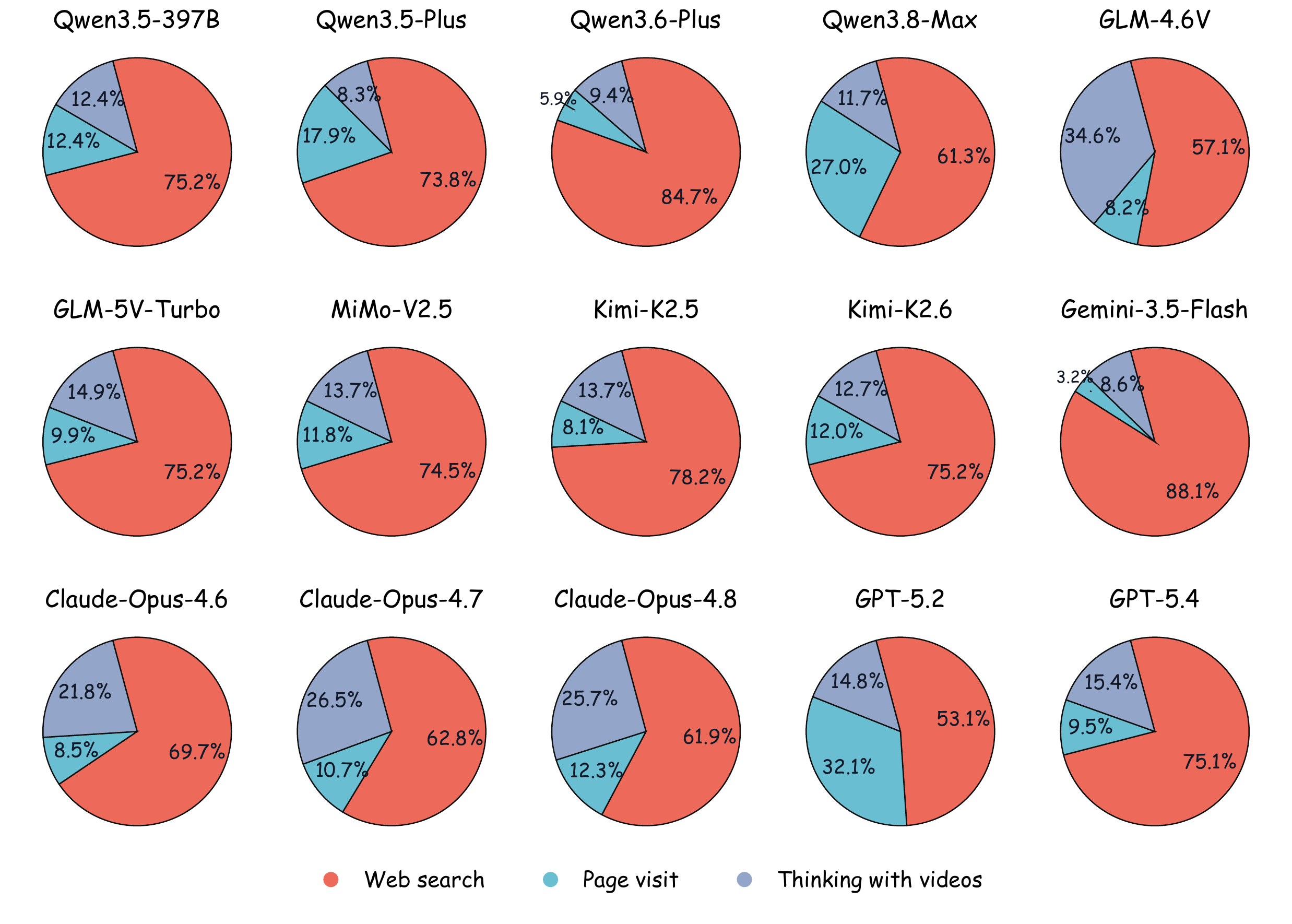}
    \caption{Tool-use distributions of 15 frontier models under the agent loop setting. Percentages are computed over all explicit tool calls on both correct and incorrect trajectories.}
    \label{fig:tool_analysis_15}
\end{figure*}

\begin{figure*}[t]
    \centering
    \includegraphics[width=\linewidth]{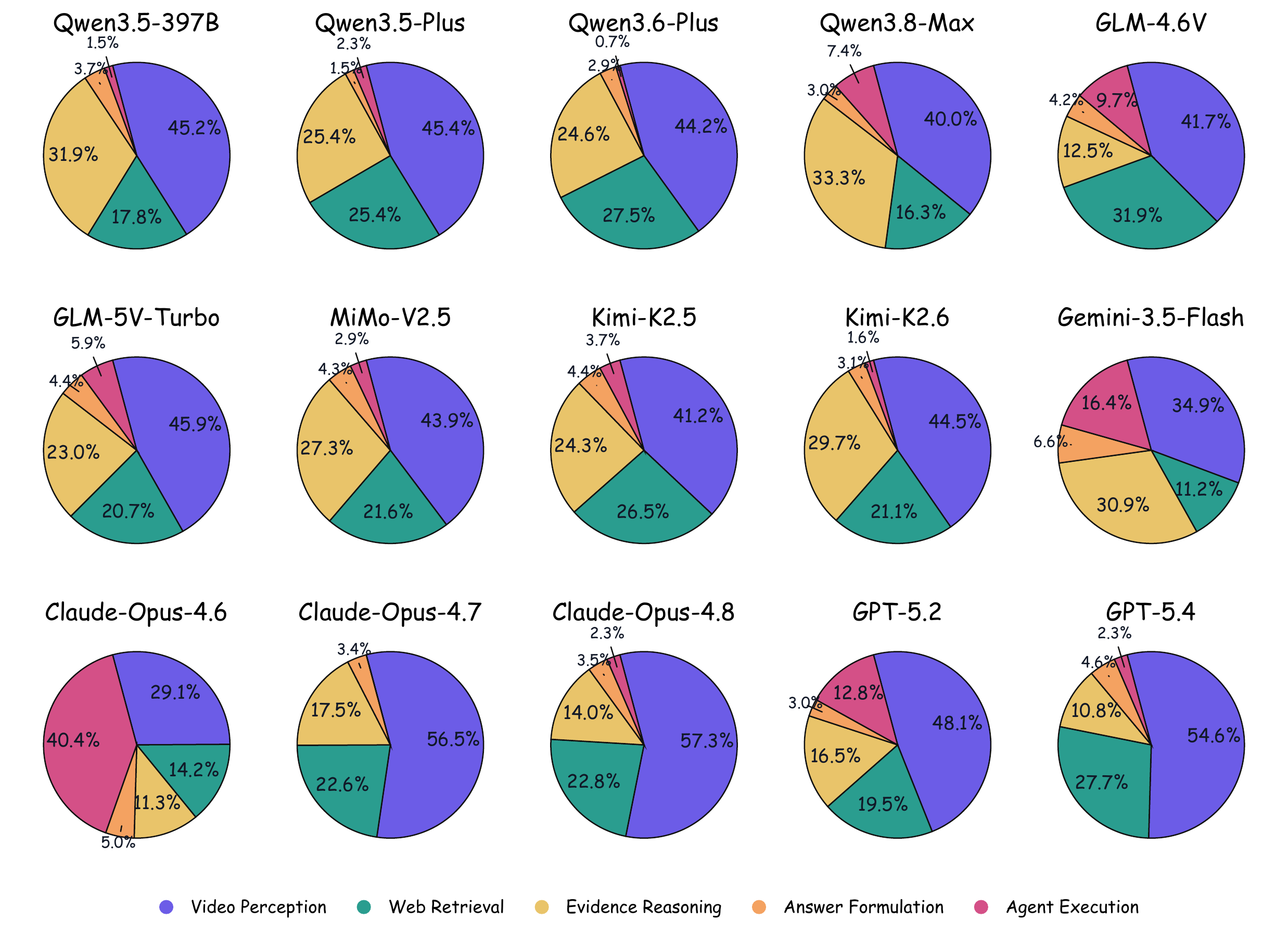}
    \caption{Distribution of dominant error causes among incorrect predictions from 15 frontier models. Each failed trajectory is assigned exactly one category.}
    \label{fig:error_analysis_15}
\end{figure*}

% \begin{figure*}[!p]
%     \centering

%     \includegraphics[
%         width=\linewidth,
%         keepaspectratio
%     ]{figs/tool_analysis_remaining15.pdf}
%     \caption{Tool-use distributions of 15 frontier models under the agent loop setting. Percentages are computed over all explicit tool calls on both correct and incorrect trajectories.}
%     \label{fig:tool_analysis_15}

%     \vspace{2em}

%     \includegraphics[
%         width=\linewidth,
%         keepaspectratio
%     ]{figs/error_analysis_remaining15.pdf}
%     \caption{Distribution of dominant error causes among incorrect predictions from 15 frontier models. Each failed trajectory is assigned exactly one category.}
%     \label{fig:error_analysis_15}
% \end{figure*}

\subsection{Additional Visualization}

As shown in Figure~\ref{fig:radar}, we provide category-level performance comparison for all 20 evaluated frontier video agents. The markedly different polygon shapes show that no model dominates uniformly across the six domains. Seed2.0-Pro \cite{seed2026seed2} is particularly strong in Geography and Technology, Kimi-K3 \cite{kimik3} performs best in Daily Life, and GPT-5.4 \cite{openaigpt54} leads in Culture and History. The radar plots also reveal substantial within-family variation, reinforcing that aggregate accuracy can conceal distinct strengths and weaknesses.

Figures~\ref{fig:tool_analysis_15} and~\ref{fig:error_analysis_15} extend the main-text analyses to the remaining 15 models and therefore complete the coverage of all evaluated systems. Web search remains the largest tool category for every additional model, accounting for 53.1\%--88.1\% of calls, although the relative use of page visits and targeted video inspection varies considerably. The error distributions are similarly consistent with the previous analysis for five representative models. Video perception is the largest dominant cause for all 15 models, comprising 34.9\%--57.3\% of their incorrect trajectories, while web retrieval and evidence reasoning explain much of the remainder. Answer-formulation errors are generally limited, whereas agent-execution failures become substantial for a small number of models. Together, these visualizations indicate that robust performance depends on grounding the correct visual anchor early, retrieving discriminative evidence, and maintaining a reliable action protocol across turns.

\begin{figure*}[t]
    \centering
    \includegraphics[width=\linewidth]{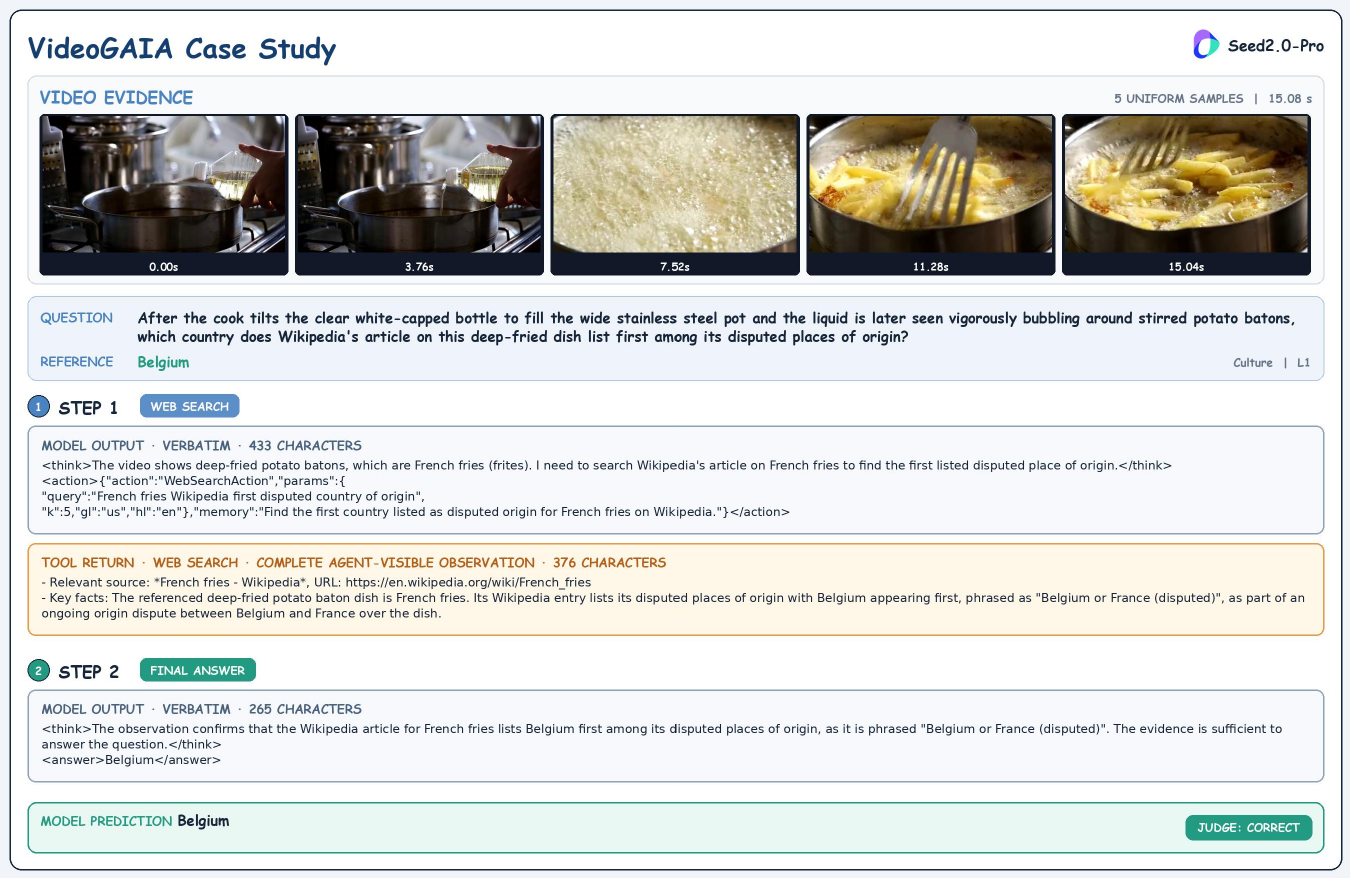}
    \caption{An example from the Culture domain of VideoGAIA.}
    \label{fig:case_culture}
\end{figure*}

\begin{figure*}[t]
    \centering
    \includegraphics[width=\linewidth]{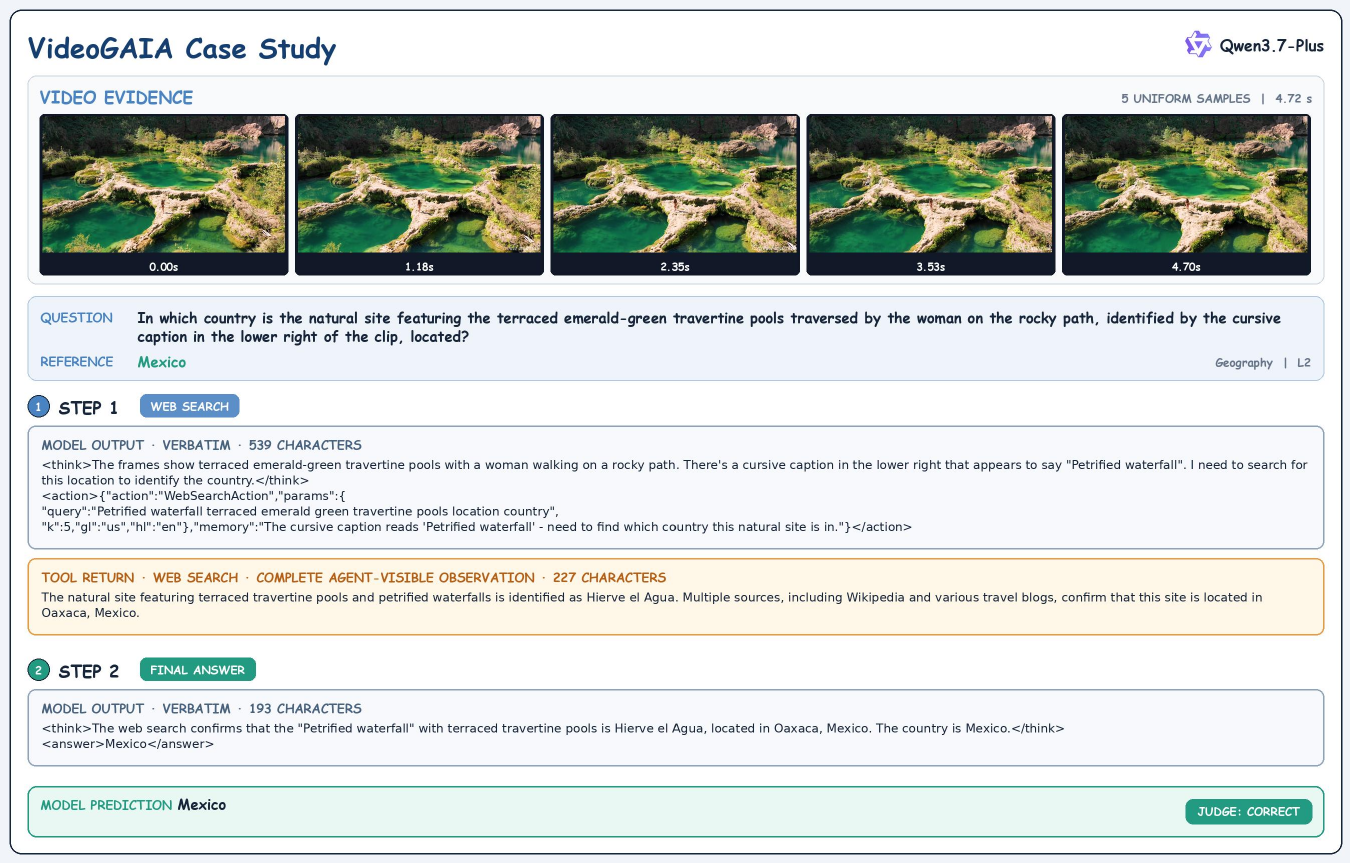}
    \caption{An example from the Geography domain of VideoGAIA.}
    \label{fig:case_geography}
\end{figure*}

% \begin{figure*}[!p]
%     \centering
%     \includegraphics[width=\linewidth,keepaspectratio]{figs/case_daily_life.pdf}
%     \caption{An example from the Daily Life domain of VideoGAIA.}
%     \label{fig:case_daily_life}
% \end{figure*}
% \clearpage

% \begin{figure*}[!p]
%     \centering
%     \includegraphics[width=\linewidth,keepaspectratio]{figs/case_history.pdf}
%     \caption{An example from the History domain of VideoGAIA.}
%     \label{fig:case_history}
% \end{figure*}
% \clearpage

% \begin{figure*}[!p]
%     \centering
%     \includegraphics[width=\linewidth,keepaspectratio]{figs/case_technology.pdf}
%     \caption{An example from the Technology domain of VideoGAIA.}
%     \label{fig:case_technology}
% \end{figure*}
% \clearpage

% \begin{figure*}[!p]
%     \centering
%     \includegraphics[width=\linewidth,keepaspectratio]{figs/case_economics.pdf}
%     \caption{An example from the Economics domain of VideoGAIA.}
%     \label{fig:case_economics}
% \end{figure*}
% \clearpage

\begin{figure*}[t]
    \centering
    \includegraphics[width=\linewidth]{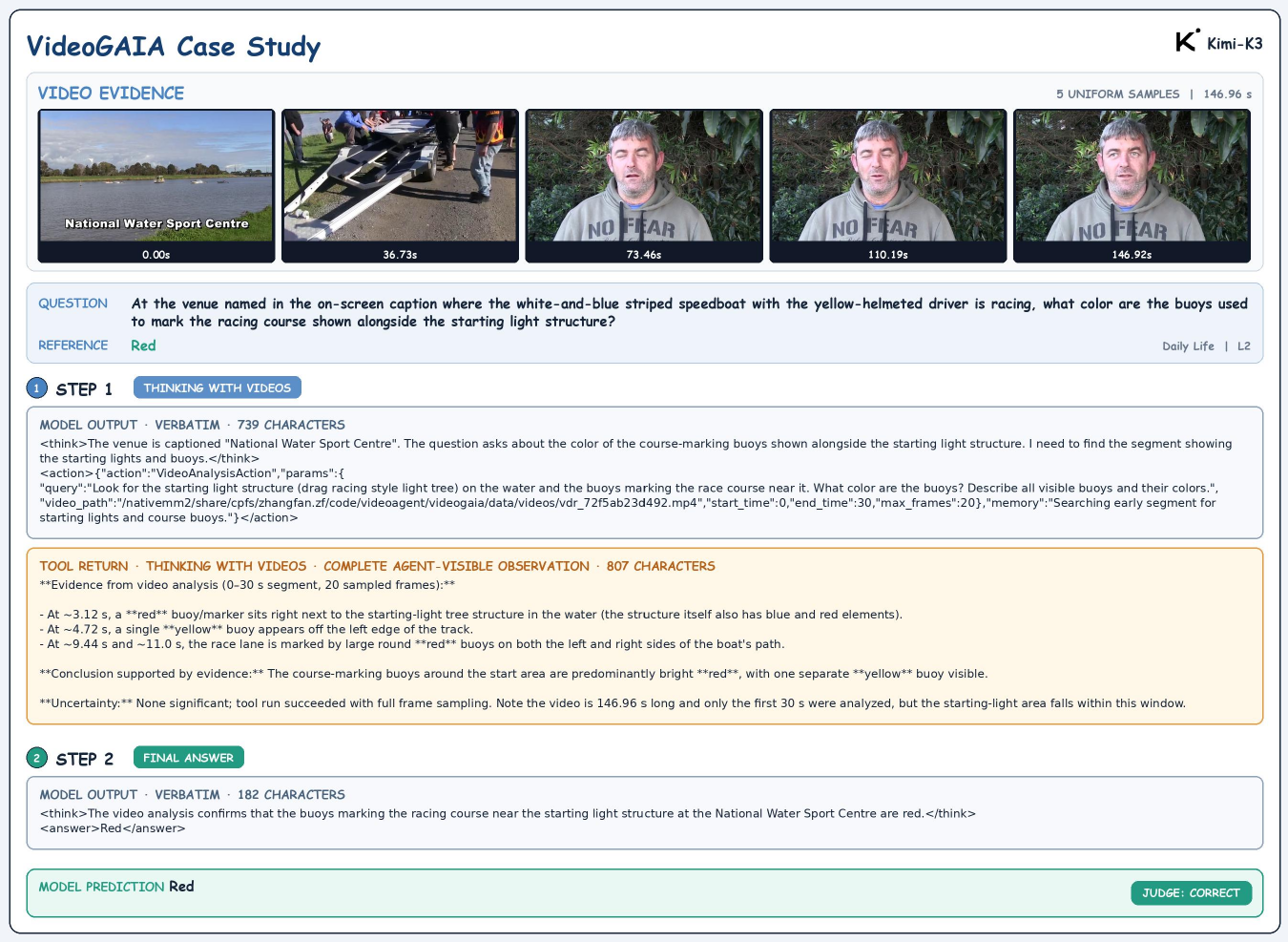}
    \caption{An example from the Daily Life domain of VideoGAIA.}
    \label{fig:case_daily_life}
\end{figure*}

\begin{figure*}[t]
    \centering
    \includegraphics[width=\linewidth]{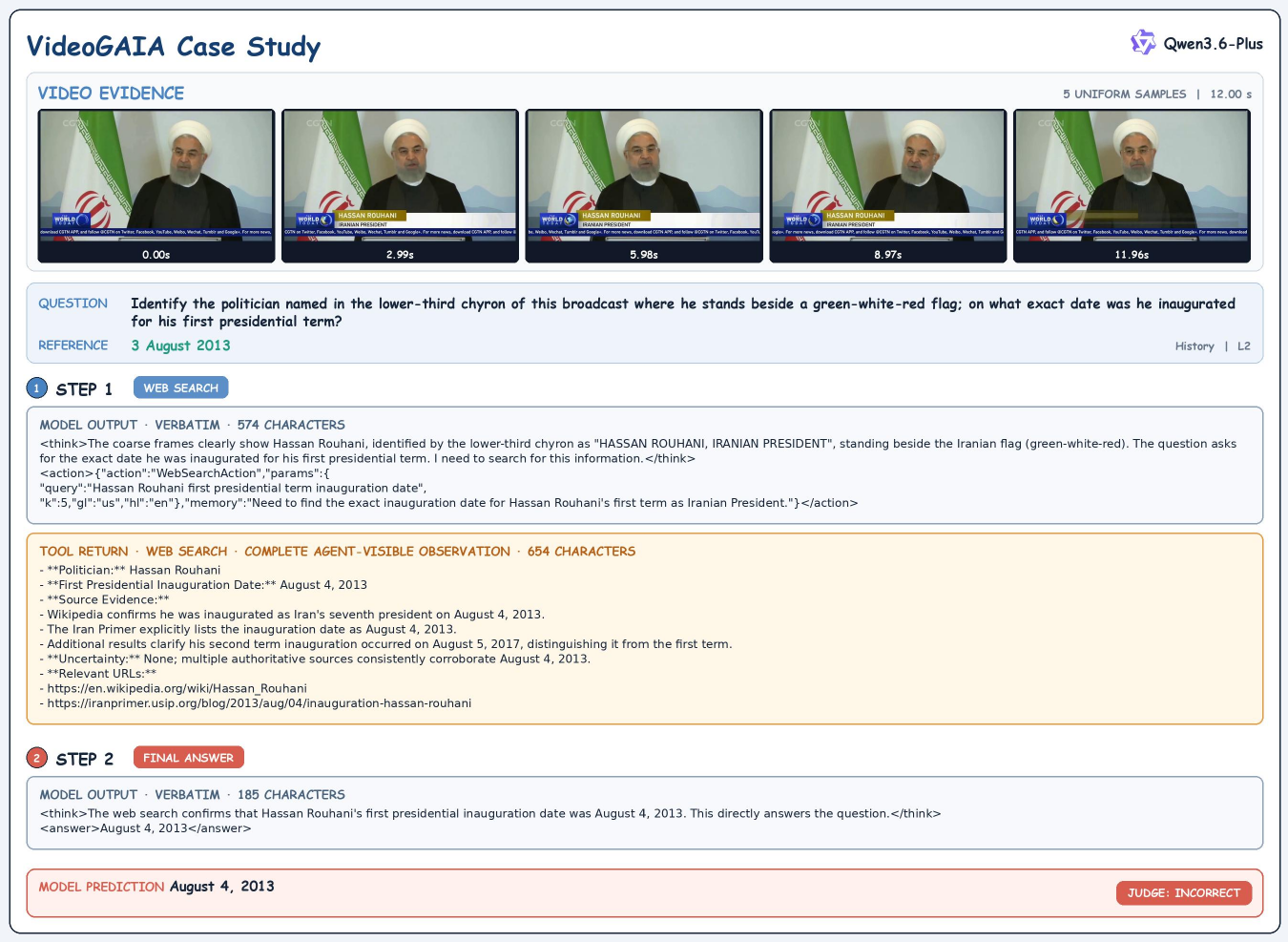}
    \caption{An example from the History domain of VideoGAIA.}
    \label{fig:case_history}
\end{figure*}

\begin{figure*}[t]
    \centering
    \includegraphics[width=\linewidth]{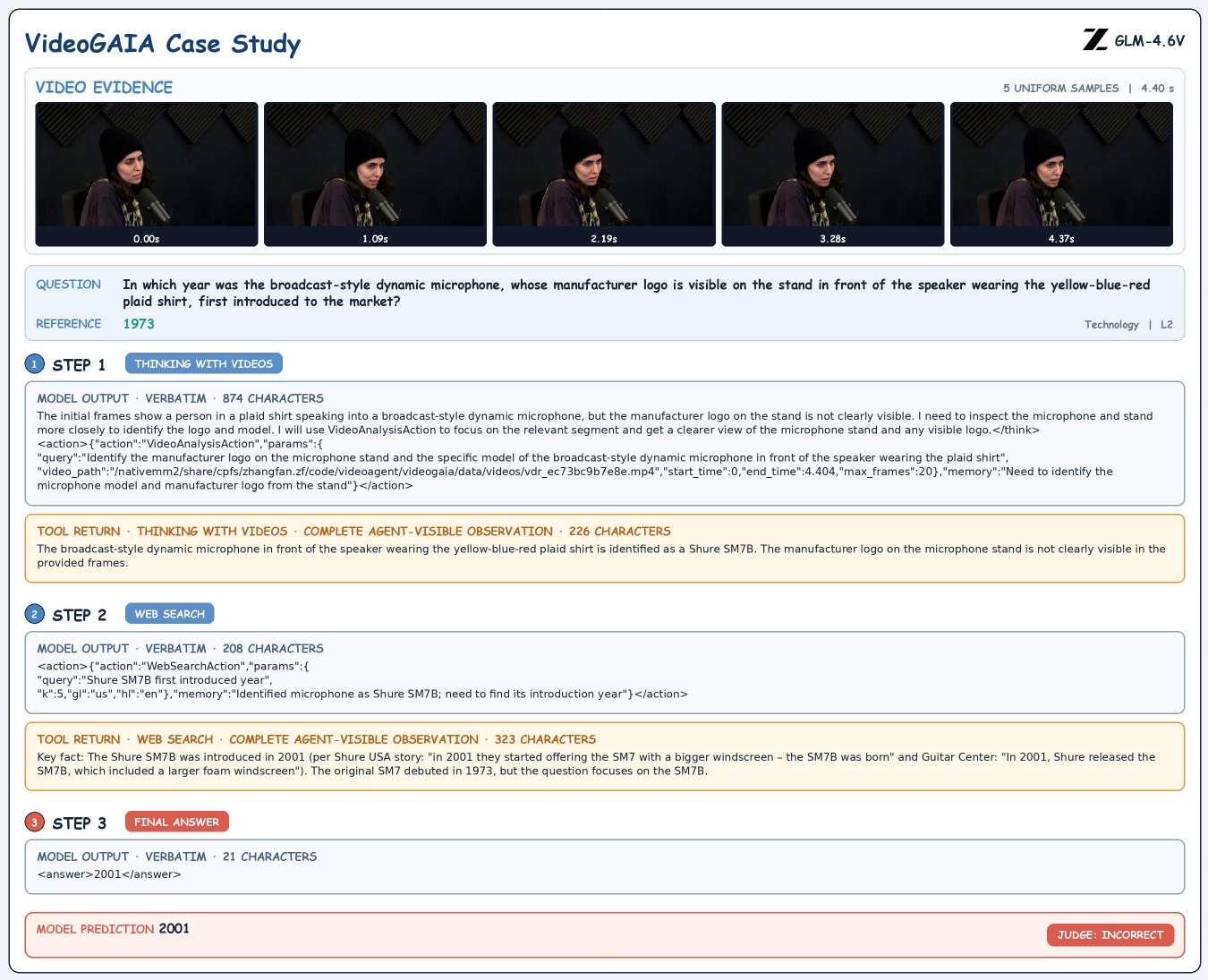}
    \caption{An example from the Technology domain of VideoGAIA.}
    \label{fig:case_technology}
\end{figure*}

\begin{figure*}[t]
    \centering
    \includegraphics[width=\linewidth]{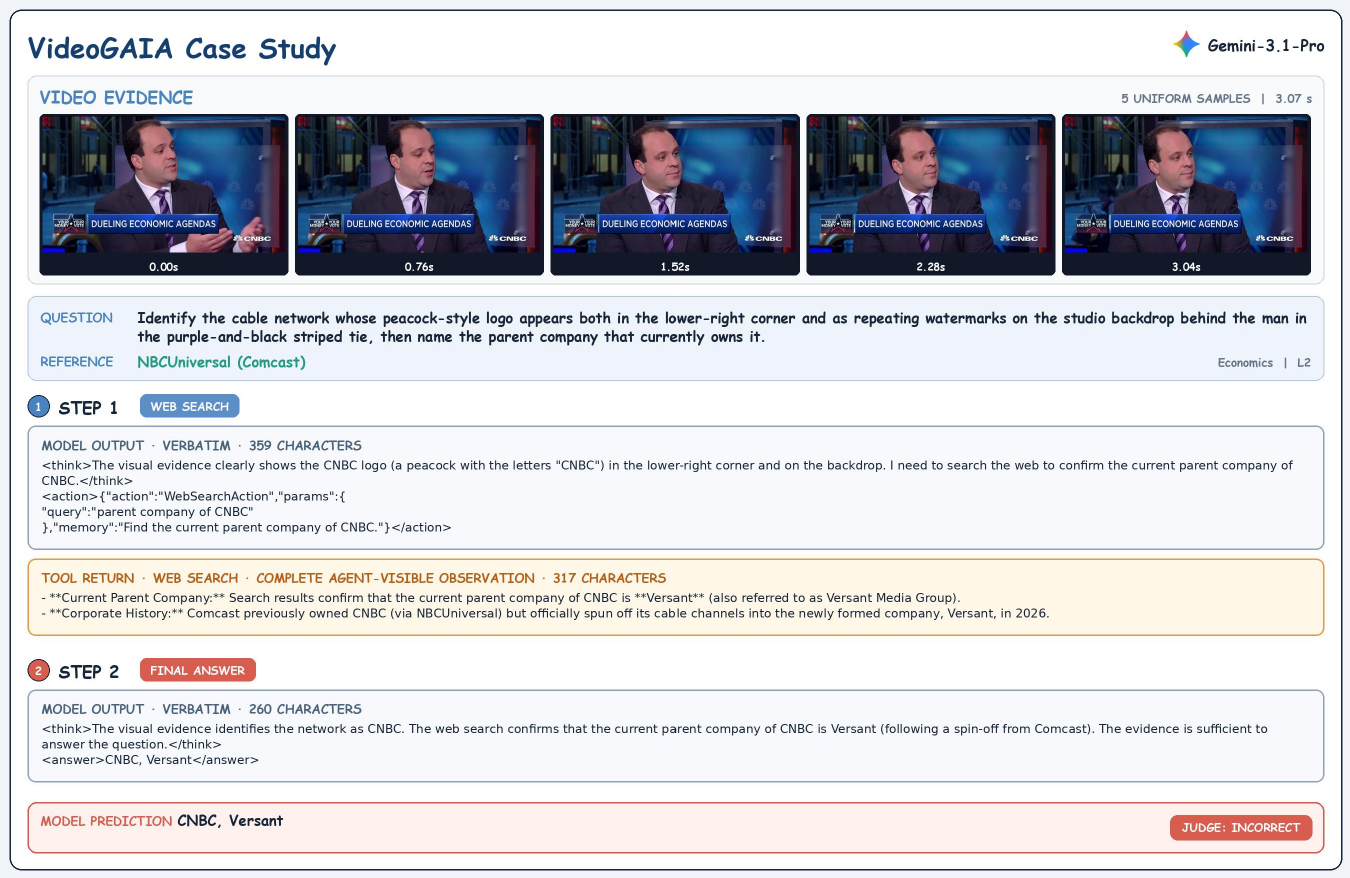}
    \caption{An example from the Economics domain of VideoGAIA.}
    \label{fig:case_economics}
\end{figure*}

\subsection{Additional Case Study}

Figures~\ref{fig:case_culture}, \ref{fig:case_geography}, \ref{fig:case_daily_life}, \ref{fig:case_history}, \ref{fig:case_technology}, and~\ref{fig:case_economics} present one complete trajectory from each VideoGAIA domain. In the successful cases, Seed2.0-Pro~\cite{seed2026seed2} and Qwen3.7-Plus~\cite{qwen37plus} ground distinctive visual clues before issuing focused web searches, while Kimi-K3~\cite{kimik3} revisits the relevant video segment to resolve the color of the racing buoys. The failed trajectories remain superficially plausible but confuse closely related facts. Qwen3.6-Plus~\cite{qwen36plus} returns August 4 rather than the annotated August 3 inauguration date, GLM-4.6V~\cite{glm46v} conflates the 2001 release of the SM7B with the original SM7 introduction in 1973, and Gemini-3.1-Pro~\cite{gemini31pro} follows a recent ownership snippet instead of the reference NBCUniversal (Comcast). Together, these examples show that tool access alone is insufficient, and reliable video agents must formulate discriminative queries, preserve uncertainty among nearby hypotheses, and cross-check retrieved facts against the exact visual and temporal context of the question.

\section{Limitations and Social Impact}

VideoGAIA has several limitations. First, the current task collection does not comprehensively cover multilingual scenarios or depend heavily on audio information. Second, to ensure consistent and controlled evaluation across models, we employ a simple ReAct-style agent loop as the video-agent harness. We do not evaluate the models within more sophisticated agent systems, such as Codex or OpenClaw, which may provide stronger context management, tool orchestration, and long-horizon planning capabilities. We leave the extension to broader modalities, languages, and advanced agent harnesses for future work. Nevertheless, these limitations do not undermine the central contribution of VideoGAIA in establishing and advancing the paradigm of agentic video understanding.

VideoGAIA has the potential to benefit research in both video understanding and multimodal agents. We hope that it will draw greater attention to video agents and encourage the community to move beyond multimodal agents that primarily operate over static images and text. In the real world, visual information is inherently continuous, dynamic, and temporally evolving. Developing agents capable of perceiving such information, seeking complementary evidence, and acting upon it is therefore essential for transforming multimodal agents from conversational systems into practical productivity tools. To date, we have not identified any potential negative societal impacts.